\documentclass[journal]{IEEEtran}

\usepackage{times}
\usepackage{epsfig}
\usepackage{graphicx}
\usepackage{amsmath}
\usepackage{amssymb}
\usepackage{cite}
\usepackage{algorithmicx,algorithm}
\usepackage[noend]{algpseudocode}
\usepackage{array, bm, placeins, lipsum}
\usepackage{booktabs, color, enumerate, microtype, multirow, subcaption}
\usepackage[x11names,dvipsnames,table]{xcolor} % use for color
\usepackage{arydshln}
\usepackage{verbatim}
\usepackage{pifont}
\usepackage{color}
\usepackage{makecell}
\newcolumntype{x}[1]{>{\centering\arraybackslash\hspace{0pt}}p{#1}}

\newcommand{\PreserveBackslash}[1]{\let\temp=\\#1\let\\=\temp}
\newcolumntype{C}[1]{>{\PreserveBackslash\centering}p{#1}}
\newcolumntype{R}[1]{>{\PreserveBackslash\raggedleft}p{#1}}
\newcolumntype{L}[1]{>{\PreserveBackslash\raggedright}p{#1}}

\usepackage{siunitx}
\ifCLASSINFOpdf
\else
\fi
\begin{document}
\title{Attribute Restoration Framework for \\ Anomaly Detection}

\author{Fei~Ye*,
        Chaoqin~Huang*,
        Jinkun~Cao,
        Maosen~Li,
        Ya Zhang,
        and Cewu Lu% <-this % stops a space
\IEEEcompsocitemizethanks{\IEEEcompsocthanksitem F. Ye, C. Huang, M. Li and Y. Zhang are with the Cooperative Medianet Innovation Center and the Shanghai Key Laboratory of Multimedia Processing and Transmissions, Shanghai
Jiao Tong University, Shanghai 200240, China. (E-mail: \{yf3310, huangchaoqin, maosen\_li, ya\_zhang\}@sjtu.edu.cn).
\IEEEcompsocthanksitem J. Cao and C. Lu are with the SJTU Machine Vision and Intelligence Group, Department of Computer Science, Shanghai Jiao Tong University, Shanghai 200240, China. (E-mail: \{caojinkun, lucewu\}@sjtu.edu.cn).}
}

\markboth{Journal of \LaTeX\ Class Files, April~2020}%
{Shell \MakeLowercase{\textit{\emph{et al.}\emph}}: Bare Demo of IEEEtran.cls for IEEE Journals}

\maketitle

\let\thefootnote\relax\footnotetext{*: contribute equally}

\begin{abstract}
With the recent advances in deep neural networks, anomaly detection in multimedia has received much attention in the computer vision community. 
While reconstruction-based methods have recently shown great promise for anomaly detection, the information equivalence among input and supervision for reconstruction tasks can not effectively force the network to learn semantic feature embeddings. We here propose to break this equivalence by erasing selected attributes from the original data and reformulate it as a restoration task, where the normal and the anomalous data are expected to be distinguishable based on restoration errors. 
Through forcing the network to restore the original image, the semantic feature embeddings related to the erased attributes are learned by the network. 
During testing phases, because anomalous data are restored with the attribute learned from the normal data, the restoration error is expected to be large.
Extensive experiments have demonstrated that the proposed method significantly outperforms several state-of-the-arts on multiple benchmark datasets, especially on ImageNet, increasing the AUROC of the top-performing baseline by 10.1\%. We also evaluate our method on a real-world anomaly detection dataset MVTec AD and a video anomaly detection dataset ShanghaiTech.  
%Source code will be made publicly available.
\end{abstract}


\IEEEpeerreviewmaketitle

\section{Introduction}
\IEEEPARstart{A}{nomaly} detection, with broad application in network intrusion detection, credit card fraud detection, and numerous other fields~\cite{chandola2009anomaly}, has received significant attention among the machine learning community. 
With the recent advances in deep neural networks, there is a heated topic on anomaly detection in multimedia, \emph{e.g.}, medical diagnosis, defect detection and intrusion detection. In this paper, we focus on anomaly detection of still images.
Anomaly detection is a technique used to identify unusual patterns that do not conform to expected behavior.
Considering the scarcity and diversity of anomalous data, anomaly detection is usually modeled as a self-supervised learning or one-class classification problem~\cite{ruff2018deep}, \emph{i.e.}, the training dataset contains only normal data and the anomalous data is not available during training.

Reconstruction-based methods~\cite{schlegl2017unsupervised, Akcay2018,Sabokrou2018Adversarially} have recently shown great promise for anomaly detection. 
Autoencoder~\cite{masci2011stacked} is adopted by most reconstruction-based methods, assuming that normal and anomalous samples could lead to significantly different embeddings and thus differences in the corresponding reconstruction errors can be leveraged to differentiate the two types of samples~\cite{Sakurada2014Anomaly}. 
\begin{figure*}[t]
\centering
\includegraphics[width=18cm]{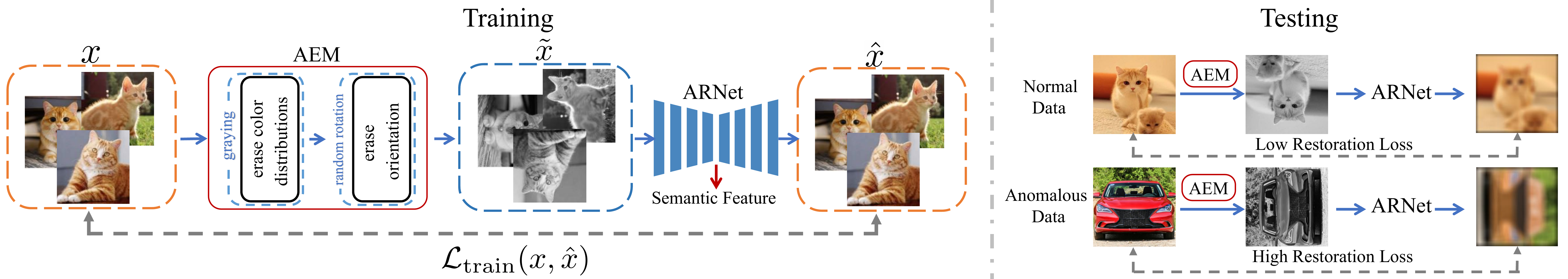}
\caption{Overview of the attribute restoration framework. During the training phase, to restore the original image, ARNet is forced to learn semantic feature embeddings related to the erased attributes. During the testing phase, the wrong restored attributes caused by the unseen semantic features will enlarge the restoration loss (the car is restored with wrong color and orientation).}
\label{fig:arnet}
\end{figure*}
However, this assumption may fail for datasets with more complex texture and structure information~\cite{russakovsky2015imagenet}. The MSE loss is shown to forces autoencoders to focus on reducing low-level pixel-wise error insensitive to human perception, rather than learning semantic features~\cite{SimilarityMetricAutoencoding,dosovitskiy2016generating}. As data complexity grows, the extracted low-level features are more likely to be shared between normal and anomalous data, leading to mixed feature embeddings. Under this situation, both normal and anomalous data could be reconstructed properly~\cite{gong2019memorizing,zong2018deep}.
To tackle this problem, various attempts have been made to introduce more efficient loss functions rather than the pixel-wise MSE loss. Adversarial training is introduced by adding a discriminator after autoencoders to judge whether its original or reconstructed image~\cite{Sabokrou2018Adversarially,deecke2018image}. Akcay \emph{et al.}~\cite{Akcay2018} adds an extra encoder after autoencoders and leverages an extra MSE loss between the two different embeddings. Despite much progress made along this line, the improvement remains limited, especially for complex datasets. Recent works~\cite{wang2019effective,bergmann2020uninformed} have shown that reconstruction-based methods fail to extract semantic features effectively. This problem may be attributed to the ``information equivalence'', i.e. the equivalence between the input and target data, which more likely leads to simple compression of the image rather than learning a semantically meaningful representation~\cite{pathak2016context}.

To achieve effective supervision and learn semantic feature embeddings, we start with breaking the information equivalence. Taking the original data as supervision, the input data is obtained by erasing selected information from the original data, creating a gap in information between the input data and the target supervision. To restore input data to the original data, the network is then forced to learn what is erased and how to restore it. The corresponding architecture is shown in Figure~\ref{fig:arnet}. In this way, we convert this task from reconstruction into restoration, a type of self-supervised learning tasks potentially effective for extracting semantic features ~\cite{pathak2016context, Jenni_2018_CVPR, zhang2016colorful}.
The design of the information erasing module is critical. If the erased information is very localized and low-level, e.g the Gaussian noise introduced in denoising autoencoders~\cite{vincent2008extracting}, it cannot effectively force the model to extract semantic features~\cite{pathak2016context}. Different from previous approaches, we introduce an Attribute Erasing Module (AEM) to remove certain attributes associated with compact semantic representations (\emph{e.g.}  color and orientation)~\cite{zhang2016colorful,long2017towards}. Since the restoration requires deeper semantic understandings of the images, the Attribute Restoration Network (ARNet) can effectively extract semantic features and the AEM controls the feature embeddings by erasing the corresponding information. Besides extracting powerful semantic features, ARNet also benefits from a unique mechanism for image restoration tasks.
Normal data can be restored properly as the erased attributes and the embedded features by the restoration network matches, which is satisfied through the training process. 
However, this match is broken when normal data and anomalous data are different regarding to the erased attribute. In this case, anomalous data can not be restored properly and suffers from high restoration errors. 

To validate the effectiveness of ARNet, we conduct extensive experiments with several benchmarks and compare them with state-of-the-art methods. Our experimental results have shown that ARNet outperforms state-of-the-art methods in terms of model accuracy and model stability for different tasks. To further evaluate with more challenging tasks, we experiment with the large-scale dataset ImageNet~\cite{russakovsky2015imagenet} and show that ARNet improves the AUROC of the top-performing baseline by 10.1\%. To illustrate that ARNet is adaptable to complex real-world environments, we experiment on a real-world anomaly detection dataset MVTec AD~\cite{bergmann2019mvtec}. We further experiment on a distorted datasets CIFAR-10-C~\cite{hendrycks2018benchmarking}. The result shows that ARNet is robust when facing low-level corruption and remains effective while other reconstruction-based methods fail, indicating that the ARNet is focused on semantic features. We also conduct T-SNE~\cite{maaten2008visualizing} visualization of latent spaces to illustrate that ARNet can extract distinctive semantic features. To the best of our knowledge, we are the first to apply image restoration to the anomaly detection problem and show impressive performance.

\section{Related Works}
\subsection{Anomaly Detection}
Depend on different tasks, anomaly detection can be roughly divided into two classes: anomaly detection in video~\cite{Kiran2018An, chu2018sparse, xu2018anomaly, xu2019video, sabokrou2016video, sabokrou2018avid, sabokrou2020deep, sabokrou2018deep, sabokrou2017deep, Sabokrou2018Adversarially} and anomaly detection in still images.
In this paper, we focus on anomaly detection in still images. Suffering from the scarcity and diversity of anomalous data, the vital challenge of anomaly detection is that the training dataset contains only normal data, leading to a lack of supervision. Judging by whether the model can be directly used in anomaly detection, popular methods can be concluded into two types accordingly: one-class classification based approaches and surrogate supervision based approaches. 

\noindent\textbf{One-class classification based approaches:}  To distinguish the anomalous data from normal data, some previous conventional methods~\cite{Eskin2000Anomaly, Yamanishi2000On, Rahmani2017Coherence, Xu2012Robust} tended to depict the normal data with statistical approaches. Through training, a distribution function was forced to fit on the features extracted from the normal data to represent them in a shared latent space. During testing, samples mapped to different statistical representations are considered as anomalous.

Some approaches tackled the anomaly detection problem by finding a hyperplane to separate normal data in the latent space.
In OC-SVM~\cite{scholkopf2001estimating}, the normal samples are mapped to the high-dimensional feature space through kernel function to get better aggregated. In the feature space, the coordinate origin is considered as the only anomalous data. Then a maximum margin hyperplane is found in feature space to better separate the mapped data from the origin.
To better aggregate the mapped data in latent space, Ruff \emph{et al.}~\cite{ruff2018deep} optimized the neural network by minimizing the volume of a hyper-sphere which encloses the network representations of the data.

Other researchers tried to find the hyperplane through generating or introducing extra anomalous data~\cite{lee2017training,hendrycks2018deep}. Lee \emph{et al.}~\cite{lee2017training} used Kullback-Leibler (KL) divergence to guide GAN to generate anomalous data closer to normal data, leading to a better training set for the classification method. 
Hendrycks \emph{et al.}~\cite{hendrycks2018deep} introduced extra data to build a multi-class classification task. The experiment revealed that even though the extra data was in limited quantities and weakly correlated to the normal data, the learned hyperplane was still effective in separating normal data.

\noindent\textbf{Surrogate supervision based approaches:} 
Many approaches modeled anomaly detection as an unsupervised learning problem and remedy the lack of supervision by introducing surrogate supervision. The model was trained to optimize the surrogate task-based objective function firstly. Then normal data can be separated with the assumption that anomalous data will result differently in the surrogate task.

Reconstruction~\cite{an2015variational,xia2015learning,schlegl2017unsupervised,zong2018deep,deecke2018image} is the most popular surrogate supervision. Based on autoencoders or variation autoencoders, this kind of method compressed normal samples into a lower-dimensional latent space and then reconstructed them to approximate the original input data. It assumed that anomalous samples would be distinguished through relatively high reconstruction errors compared with normal samples. 
Sakurada \emph{et al.}~\cite{Sakurada2014Anomaly} were the first to apply the autoencoder to anomaly detection. This work further indicated that the learned features in the hidden layer of autoencoders were distinguishable between normal and anomalous data. Based on that, Nicolau \emph{et al.}~\cite{nicolau2016hybrid} introduced density estimation to estimate the different distribution in the latent space of autoencoder. It assumed that anomalous data would hold lower density in latent space.
Some recent works~\cite{SimilarityMetricAutoencoding,dosovitskiy2016generating} indicated mean square error (MSE) loss function, adopted by most reconstruction-based methods, forces the network to focus on pixel-wise error rather than learning semantic features. When dealing with more complex data, the learned low-level features are more likely to be shared and lead to good reconstruction results in both normal and anomalous data. 

To tackle this problem, some recent approaches continued to follow the reconstruction based method by introducing more efficient loss function rather than MSE. Adversarial training is employed to optimize the autoencoder and its discriminator is leveraged to further enlarge the reconstruction error gap between normal and anomalous data~\cite{Sabokrou2018Adversarially,sabokrou2020deep}. To make the method more robust against noises, Gaussian noise is added to the input training samples and then fed to the encoder. To detect the irregularity in videos, Sabokrou \emph{et al.}~\cite{sabokrou2018avid} proposed an architecture to detect and localize the irregularity simultaneously. Training by the normal data only, the model learns to replace the irregularity in the video with a dominant concept, in which this process is noted as image inpainting. 
Based on~\cite{Sabokrou2018Adversarially}, Akcay \emph{et al.}~\cite{akccay2019skip} leveraged another encoder to embed the reconstruction results to the subspace where to calculate the reconstruction error. Similarly, Wang \emph{et al.}~\cite{wang2019advae} employed adversarial training under a variational autoencoder framework with the assumption that normal and anomalous data follows different Gaussian distribution. 
Zenati \emph{et al.}~\cite{zenati2018efficient} trained a BiGAN model and employed a discriminator to add supervision to encoder and decoder simultaneously.
Gong \emph{et al.}~\cite{gong2019memorizing} augmented the autoencoder with a memory module and developed an improved autoencoder called memory-augmented autoencoder to strengthen reconstructed errors on anomalies. Perera \emph{et al.}~\cite{perera2019ocgan} applied two adversarial discriminators and a classifier on a denoising autoencoder. By adding constraint and forcing each randomly drawn latent code to reconstruct examples like the normal data, it obtained high reconstruction errors for the anomalous data.

Other approaches tackled this problem by introducing new self-supervised methods as surrogate tasks. Golan \emph{et al.}~\cite{golan2018deep} introduced a self-supervised method \cite{gidaris2018unsupervised} into anomaly detection by applying dozens of image geometric transforms and created a self-labeled dataset for transformation classification, assuming that the transformation of anomalous data can not be classified properly. Wang \emph{et al.}~\cite{wang2019effective} further introduced a self-supervised method~\cite{noroozi2016unsupervised} into anomaly detection by applying Jigsaw puzzles to extend the above self-labeled dataset.

\subsection{Self-supervised Learning Methods}
Self-supervised representation learning leverages input data itself as supervision and benefits almost all types of downstream tasks like classification, detection and segmentation. 
Examples of self-supervised tasks are recognizing the geometric transformation applied to an image~\cite{gidaris2018unsupervised}, 
predicting the relative position between a pair of random patches from an image~\cite{doersch2015unsupervised}, solving Jigsaw puzzles after randomly swapping the image patching~\cite{noroozi2016unsupervised}. Some works tackled the self-supervised learning problem through restoration. It assumed that by restoring the damaged image, the network is forced to learn robust feature embeddings.
Denoising autoencoders~\cite{vincent2008extracting} add alternative corrupting noises to the original data, and require the autoencoder network to undo the damage. Pathak \emph{et al.}~\cite{pathak2016context} indicated that the damaged region should have a relatively large size, otherwise the model could restore the damaged image through the local non-semantic feature. Pathak \emph{et al.}~\cite{pathak2016context} randomly blanked out a region from the original image and employed an autoencoder to restore.
Jenni \emph{et al.}~\cite{Jenni_2018_CVPR} tackle the problem by removing and inpainting information at a more abstract level (the internal representation), rather than at the raw data level. Denton~\cite{denton2016semi} indicated that previous work was difficult to generate large image patches that look realistic. To address this, a low resolution but intact version of the original image was extra fed to the network to guide reconstruction. To be noted that in the restoration framework, the information erasing module is non-parameterized and separated from the network.

In this paper, we introduce the restoration framework for anomaly detection to control feature embedding and extract semantic features. The information erasing module is different from traditional restoration-based unsupervised learning in this paper. In ARNet, a novel and effective attribute erase module is utilized to erase certain object attribute from the image.  In anomaly detection, there are several approaches related to our work. Sabokrou \emph{et al.}~\cite{Sabokrou2018Adversarially,sabokrou2020deep} applying Denoising autoencoders\cite{vincent2008extracting}, which can also be considered as restoration-based anomaly detection. Sabokrou \emph{et al.}~\cite{sabokrou2018avid} indicated that AVID operates similarly to a denoising network, which replaces the irregularity in the video with a dominant concept, it can be considered as inpainting-based anomaly detection. However, Pathak \emph{et al.}~\cite{pathak2016context} indicated that the Gaussian noise introduced in Denoising autoencoders is typically very localized and low-level, thus does not require much semantic information to restore. Different from these approaches, by forcing the network to restore the erased attributes, ARNet can effectively control feature embedding and extract semantic features.

\section{Attribute Restoration Framework}
\begin{figure}[t]
\centering
\includegraphics[width=7.5cm]{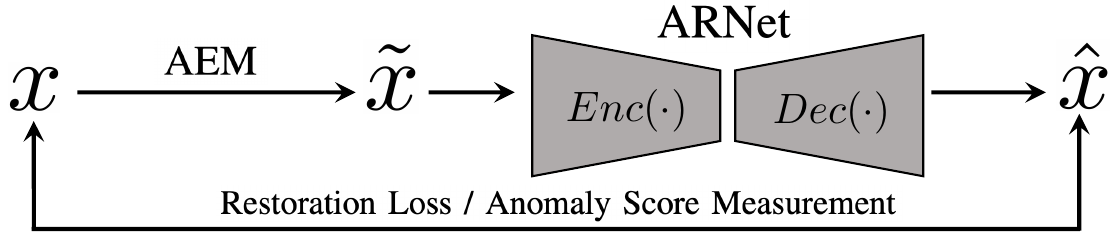}
\caption{Pipeline for anomaly detection with attribute restoration framework with mathematical expression.}
\label{img:ARNet}
\end{figure}

We first formulate the problem of anomaly detection. 
Let $\mathcal{X}$, $\mathcal{X_\mathrm{n}}$, and $\mathcal{X_\mathrm{an}}$ denote the sets of entire dataset, normal dataset and anomalous dataset, respectively, where $\mathcal{X_\mathrm{n}} \cup \mathcal{X_\mathrm{an}} = \mathcal{X} $ and $\mathcal{X_\mathrm{n}} \cap \mathcal{X_\mathrm{an}} = \emptyset $. Given any image $x \in \mathcal{X}$, where $x \in \mathbb{R}^{C \times H \times W}$, and $C$, $H$ and $W$ denote the dimensions of image channels, height and width, the goal is to build a model $\mathcal{M}(\cdot)$ for discriminating whether $x \in \mathcal{X_\mathrm{n}} $ or $x \in \mathcal{X_\mathrm{an}}$. To solve the above problem, we propose the attribute restoration framework, which consists of three parts: (1) Attribute Erasing Module (AEM): erase certain attributes of images to create an image restoration task; (2) Attribute Restoration Network (ARNet): use the original images as supervision and the images after erasing certain attributes as inputs to train a model for restoring the images against the attribute absence; (3) Anomaly measurement: establish a link between the image restoration task and the image anomaly detection task. The corresponding structure is shown in Figure~\ref{img:ARNet}. 

\subsection{Attribute Erasing Module (AEM)}\label{method}
Attribute Erasing Module (AEM) aims to erase a set of attributes from the objects, enforcing information inequivalence between input and output data and turn the task into restoration. The effective attribute for anomaly detection should satisfy the following assumptions:
\begin{itemize}
    \item The erased attribute should be connected to the semantic information of the normal data and be shared among the normal data.
    \item The erased attribute should be either different or connected to different semantic information between normal and anomalous data. If the normal and anomalous data share the same attribute and the attribute is connected to the same semantic information, the ARNet is hard to distinguish the normal data and anomalous data through this attribute, as anomalous data can be restored properly using the shared features learning from the normal data.
    \item The attributes can be erased by a module which does not rely on extra dataset or labels; otherwise, this will require an additional training process for attribute erasing.
\end{itemize}

We take a concrete example to further reveal the details to design the Attribute Erasing Module. By human prior, the semantic information connected to the orientation is shared within a class but different between classes, e.g. the wheels of the cars are always at the bottom of the images while the circle of the digit number ``9'' is always at the top of the images, which meets the first and the second condition. To erase the orientation of these objects, we can employ a random rotation operation, which rotates the images with a randomly selected angle. This orientation erasing operation does not need to introduce an additional training process, which also meets the third condition we discussed above. 

The main challenge of unsupervised anomaly detection is that anomalous data is not available during training, leaving no guarantee that the second assumption is satisfied by all kinds of anomalous data. Fortunately, as the chosen attribute is connected to the semantic information, it is hard to be shared between normal and anomalous data (see Section~\ref{sec:attribute share} for more discussion). In addition, although some attributes cannot be used to distinguish between normal and anomalous data when the second condition is not satisfied, it only causes the anomaly detection performance of the image restoration task degrading to that of the image reconstruction task. To alleviate this problem, we propose a set of attribute erase operations to increase the probability that at least one attribute could meet the second condition.
The Attribute Erasing Module works as follows:
Suppose we have a set of attribute erasing operations $\mathcal{O} = \{ f_{\mathrm{O}_k}(\cdot) | k=1,\dots,K \}$, where $f_{\mathrm{O}_{k}}(\cdot)$ denotes the \emph{k}-th attribute erasing operation. Given $x_{\rm n} \in \mathcal{X_\mathrm{n}}$, the data after AEM should be $\tilde{x}_{\rm n} = F_{\mathcal{O}}(x_{\rm n}) := f_{\mathrm{O}_k}( f_{\mathrm{O}_{k-1}}( \cdots f_{\mathrm{O}_1}( x_{\rm n})))$.

\subsection{Attribute Restoration Network}
We now present the \emph{Attribute Restoration Network} (ARNet) in detail. ARNet is based on an encoder-decoder framework to restore the original images. 
In the training phase, given $\tilde{x}_{\rm n}$ after AEM,
the proposed ARNet takes the $\tilde{x}_{\rm n}$ as the inputs, and attempts to inversely restore the original training samples $x_{\rm n}$. Given $\tilde{x}_{\rm n}$, the restored sample $\hat{x}_{\rm n}$ is expressed as
\begin{equation}
\label{eq:pipeline}
\hat{x}_{\rm n}=\mathcal{M}(\tilde{x}_{\rm n})=Dec(Enc(\tilde{x}_{\rm n})),
\end{equation}                     
where $\mathcal{M}(\cdot)$ indicates the model of ARNet, while $Enc(\cdot)$ and $Dec(\cdot)$ indicate encoder and decoder of ARNet. 
Note that while ARNet is employed for the image restoration tasks, it is different from existing autoencoders in that the inputs and outputs are asymmetrical, \emph{i.e.},~ARNet needs to restore attributes erased by the Attribute Erasing Module.

To train ARNet for effective anomaly detection, a likelihood-based restoration loss is employed as loss function. $\ell_2$ loss is utilized to measure the distances between the restored samples and targets since it is smoother and distributes more punishments on the dimensions with larger generation errors. Let the target image be $x_{\rm n}$, the training loss is formulated as
\begin{equation}
\label{MC}
\begin{split}
    \mathcal{L}_{\rm train} =
    \mathbb{E}_{x_{\rm n}\sim p(\mathbf{x}_{\rm n})}
    \left \| \mathcal{M}(\tilde{x}_{\rm n})-x_{\rm n} \right \|_2^2,
\end{split}
\end{equation}
where $\|\cdot\|_2$ denotes the $\ell_2$ norm and $p({\bf x}_{\rm n})$ indicates the distribution of normal data. To approximate the expectation operation in Eq.~\eqref{MC}, in each mini-batch, we randomly select a sample ${x}_{\rm n}$ and obtain $\tilde{x}_{\rm n} = F_{\mathcal{O}}(x_{\rm n})$. Then we calculate the average cost between any $x_n$ and corresponding $\mathcal{M}(\tilde{x}_{\rm n})$.

\subsection{Anomaly Measurement}\label{sec:score}
To establish a link between the image restoration task and the image anomaly detection task, in the test phase, we design a metric based on the restoration error to distinguish whether one sample belongs to the normal set. Both normal and anomalous data are fed into the model, which are utilized together to determine whether a query sample is anomalous. 
In the test phase, we calculate the restoration error of each input image $x$ for anomaly detection. We suppose that the restorations of normal samples show much smaller errors than the anomalous samples due to the specific image restoration scheme. We note that $\ell_1$ loss is more suitable to measure the distance between outputs and original images. Let the test sample be $x$, the anomaly score is formulated as
\begin{equation}
\begin{split}
{\cal S}_{\rm test}(x) = \left \| \mathcal{M}(F_{\mathcal{O}}(x))-x \right \|_1,
\end{split}
\end{equation}
where $\|\cdot\|_1$ denotes the $\ell_1$ norm. 

However, $f_{O_k}$ may function through randomization, in which the original fixed operation is reformulated as a random selection $f_{\hat{O}_k}$ from an operation set  $\{f_{\hat{O}_{k,j_k}} | j_k=1,\dots,m_k \}$ with size of $m_k$. For example, we employ random rotation to formulate the orientation erasing operation, where the rotation angle is randomly selected from a fixed set, such as several discrete angle options, $\{\ang{0}, \ang{90},\ang{180},\ang{270}\}$. Accordingly, as the $F_{\mathcal{O}}$ is the compound function of $f_{O_k}$, $F_{\mathcal{O}}$ is reformulated as $F_{\hat{\mathcal{O}}}(\cdot) = f_{\hat{\mathrm{O}}_k}( f_{\hat{\mathrm{O}}_{k-1}}( \cdots f_{\hat{\mathrm{O}}_1}(\cdot)))$, where $F_{\hat{\mathcal{O}}}(\cdot)$ is a random selection from the set $\{ F_{\hat{\mathrm{O}}_i}(\cdot) | i=1,\dots,N \}$ with size $\emph{N}=\prod_{k=1}^K m_k$.
Note that, when $m_k=1$, $f_{O_k}=f_{\hat{O}_k}$. 
During the test process, we need to traverse all selections $F_{\hat{\mathrm{O}}_i}(\cdot)$ and set average restoration error as the anomaly score, which is reformulated as
\begin{equation}
\begin{split}
{\cal S}_{\rm test}'(x) = 
\frac{1}{N}
& \sum_{i=1}^{N}
\left \| \mathcal{M}(F_{\mathcal{\hat{O}}_i}(x))-x \right \|_1.
\end{split}
\end{equation}

We notice that the restoration errors under some $F_{\hat{\mathrm{O}}_i}(\cdot)$ may larger than the others in natural since different tasks have different restoration difficulties. In this case, given the same input sample, different $F_{\hat{\mathrm{O}}_i}(\cdot)$ lead to different restoration errors and the final anomaly score may has a bias if we average these restoration errors naively. To make each $F_{\hat{\mathrm{O}}_i}(\cdot)$ contributes equally to the final anomaly score, we use the original training data and calculate the mathematical expectation of the restoration error for each $F_{\hat{\mathrm{O}}_i}(\cdot)$ as a normalization, and set the final anomaly score as
\begin{equation}
\begin{split}
{\cal S}_{\rm test}''(x) = 
\frac{1}{N}
& \sum_{i=1}^{N}
\frac{\left \| \mathcal{M}(F_{\mathcal{\hat{O}}_i}(x))-x \right \|_1}
{\mathbb{E}_{x_{\rm n}\sim p({\bf x}_{\rm n})}\left \| \mathcal{M}(F_{\mathcal{\hat{O}}_i}(x_{\rm n}))-x_{\rm n} \right \|_1 ]},
\end{split}
\end{equation}
where $p({\bf x}_{\rm n})$ indicates the distribution of normal data, as well as being consistent with the distribution of training set. A normal sample leads to a low anomaly score; the higher value $\mathcal{S}_{\rm test}''(x)$  obtained, the higher probability for the sample $x$ to be anomalous.

\subsection{Discussion: Restoration vs. Reconstruction}

Both image reconstruction and image restoration tasks can be implemented with an encoder-decoder architecture. The differences are summarized in three folds. First, different from reconstruction, the input and output for ARNet are asymmetric which is achieved with an Attribute Erasing Module.
The erased information of anomalous data may not be restored properly through feature embeddings learned from the normal data, leading to high anomaly scores for anomalous data. 
Secondly, unlike the reconstruction-based methods, especially vanilla AE, which blindly learns uncontrollable features from normal data, the restoration-based framework leverages the attribute erasing to guide the feature embedding and thus enables the embedding of semantic features. 
Thirdly, in the final anomaly detection phase, the two methods differ in the way to obtain the final anomaly scores. Different from the reconstruction-based methods, for the restoration-based framework, multiple restoration losses produced by multiple attribute erasing operations are weighted and summed to obtain the anomaly scores. These weights can be obtained from the training data, which has been discussed in Section~\ref{sec:score}.

\renewcommand \arraystretch{0.95}
\begin{table*}[!htb]
\centering
\caption{Average area under the ROC curve (AUROC) in \% of anomaly detection methods. For every dataset, each model is trained on the single class, and tested against all other classes. ``SD'' means standard deviation among classes. The best performing method is in bold.}
	\footnotesize
	\begin{minipage}[t]{0.95\textwidth}
	\begin{tabular}{cx{2.0cm}x{0.6cm}x{0.6cm}x{0.6cm}x{0.6cm}x{0.6cm}x{0.6cm}x{0.6cm}x{0.6cm}x{0.6cm}x{0.6cm}x{0.6cm}x{0.8cm}}
	\toprule
	Dataset & Method & 0 & 1 & 2 & 3 & 4 & 5 & 6 & 7 & 8 & 9 & \textbf{avg} & SD\\
	\cmidrule(lr){1-1} \cmidrule(lr){2-2} \cmidrule(lr){3-3} \cmidrule(lr){4-4} \cmidrule(lr){5-5} \cmidrule(lr){6-6} \cmidrule(lr){7-7} \cmidrule(lr){8-8} \cmidrule(lr){9-9} \cmidrule(lr){10-10} \cmidrule(lr){11-11} \cmidrule(lr){12-12} \cmidrule(lr){13-13} \cmidrule(lr){14-14}
		& VAE~\cite{kingma2013auto} 
		& 92.1 & \textbf{99.9} & 81.5 & 81.4 & 87.9 & 81.1 & 94.3 & 88.6 & 78.0 & 92.0 & 87.7 & 7.05\\
		& ALOCC~\cite{Sabokrou2018Adversarially}
		& 99.5 & 99.1 & 92.0 & 92.1 & 93.5 & 84.7 & 97.5 & 94.1 & 87.5 & 92.8 & 93.3 & 4.70\\
		& AnoGAN~\cite{schlegl2017unsupervised} 
		& 99.0 & 99.8 & 88.8 & 91.3 & 94.4 & 91.2 & 92.5 & 96.4 & 88.3 & 95.8 & 93.7 & 4.00\\
		& ADGAN~\cite{deecke2018image} 
		& 99.5 & \textbf{99.9} & 93.6 & 92.1 & 94.9 & 93.6 & 96.7 & 96.8 & 85.4 & 95.7 & 94.7 & 4.15\\
		MNIST& GANomaly~\cite{Akcay2018} 
		& 97.2 & 99.6 & 85.1 & 90.6 & 94.5 & 94.9 & 97.1 & 93.9 & 79.7 & 95.4 & 92.8 & 6.12\\
		& OCGAN~\cite{perera2019ocgan} 
		& \textbf{99.8} & \textbf{99.9} & 94.2 & 96.3 & 97.5 & 98.0 & 99.1 & 98.1 & 93.9 & 98.1 & 97.5 & 2.10\\
		& GeoTrans~\cite{golan2018deep} 
		& 98.2 & 91.6 & \textbf{99.4} & 99.0 & \textbf{99.1} & \textbf{99.6} & \textbf{99.9} & 96.3 & \textbf{97.2} & \textbf{99.2} & 98.0 & 2.50\\
		\cmidrule(lr){2-2} \cmidrule(lr){3-3} \cmidrule(lr){4-4} \cmidrule(lr){5-5} \cmidrule(lr){6-6} \cmidrule(lr){7-7} \cmidrule(lr){8-8} \cmidrule(lr){9-9} \cmidrule(lr){10-10} \cmidrule(lr){11-11} \cmidrule(lr){12-12} \cmidrule(lr){13-13} \cmidrule(lr){14-14}
		& AE & 98.8 & 99.3 & 91.7 & 88.5 & 86.2 & 85.8 & 95.4 & 94.0 & 82.3 & 96.5 & 91.9 & 5.90\\
		& OURS & 98.6 & \textbf{99.9} & 99.0 & \textbf{99.1} & 98.1 & 98.1 & 99.7 & \textbf{99.0} & 93.6 & 97.8 & \textbf{98.3} & \textbf{1.78}\\
		        \cmidrule(lr){1-14}
		& DAGMM~\cite{zhai2016deep} 
        & 42.1 & 55.1 & 50.4 & 57.0 & 26.9 & 70.5 & 48.3 & 83.5 & 49.9 & 34.0 & 51.8 & 16.47\\
		& DSEBM~\cite{zong2018deep} 
		& 91.6 & 71.8 & 88.3 & 87.3 & 85.2 & 87.1 & 73.4 & 98.1 & 86.0 & 97.1 & 86.6 & 8.61\\
		Fashion- & ADGAN~\cite{deecke2018image} 
		& 89.9 & 81.9 & 87.6 & 91.2 & 86.5 & 89.6 & 74.3 & 97.2 & 89.0 & 97.1 & 88.4 & 6.75\\
		MNIST & GANomaly~\cite{Akcay2018} 
		& 80.3 & 83.0 & 75.9 & 87.2 & 71.4 & 92.7 & 81.0 & 88.3 & 69.3 & 80.3 & 80.9 & 7.37\\
		& GeoTrans~\cite{golan2018deep} 
		& \textbf{99.4} & 97.6 & \textbf{91.1} & 89.9 & \textbf{92.1} & \textbf{93.4} & 83.3 & \textbf{98.9} & 90.8 & \textbf{99.2} & 93.5 & 5.22\\
		\cmidrule(lr){2-2} \cmidrule(lr){3-3} \cmidrule(lr){4-4} \cmidrule(lr){5-5} \cmidrule(lr){6-6} \cmidrule(lr){7-7} \cmidrule(lr){8-8} \cmidrule(lr){9-9} \cmidrule(lr){10-10} \cmidrule(lr){11-11} \cmidrule(lr){12-12} \cmidrule(lr){13-13} \cmidrule(lr){14-14}
		& AE & 71.6 & 96.9 & 72.9 & 78.5 & 82.9 & 93.1 & 66.7 & 95.4 & 70.0 & 80.7 & 80.9 & 11.03\\
		& OURS & 92.7 & \textbf{99.3} & 89.1 & \textbf{93.6} & 90.8 & 93.1 & \textbf{85.0} & 98.4 & \textbf{97.8} & 98.4 & \textbf{93.9} & \textbf{4.70}\\
        \cmidrule(lr){1-14}
	& VAE~\cite{kingma2013auto} 
	& 62.0 & 66.4 & 38.2 & 58.6 & 38.6 & 58.6 & 56.5 & 62.2 & 66.3 & 73.7 & 58.1 & 11.50\\
	& DAGMM~\cite{zhai2016deep} 
	& 41.4 & 57.1 & 53.8 & 51.2 & 52.2 & 49.3 & 64.9 & 55.3 & 51.9 & 54.2 & 53.1 & 5.95\\
	& DSEBM~\cite{zong2018deep} 
	& 56.0 & 48.3 & 61.9 & 50.1 & 73.3 & 60.5 & 68.4 & 53.3 & 73.9 & 63.6 & 60.9 & 9.10\\
	& ALOCC~\cite{Sabokrou2018Adversarially}
	& 62.0 & 71.7 & 53.7 & 56.0 & 58.7 & 56.3 & 61.2 & 60.5 & 74.4 & 67.1 & 62.2 & 6.90\\
	CIFAR- & AnoGAN~\cite{schlegl2017unsupervised} 
	& 61.0 & 56.5 & 64.8 & 52.8 & 67.0 & 59.2 & 62.5 & 57.6 & 72.3 & 58.2 & 61.2 & 5.68\\
	10& ADGAN~\cite{deecke2018image} 
	& 63.2 & 52.9 & 58.0 & 60.6 & 60.7 & 65.9 & 61.1 & 63.0 & 74.4 & 64.4 & 62.4 & 5.56\\
	& GANomaly~\cite{Akcay2018} 
	& \textbf{93.5} & 60.8 & 59.1 & 58.2 & 72.4 & 62.2 & 88.6 & 56.0 & 76.0 & 68.1 & 69.5 & 13.08\\
	& OCGAN~\cite{perera2019ocgan} 
	& 75.7 & 53.1 & 64.0 & 62.0 & 72.3 & 62.0 & 72.3 & 57.5 & 82.0 & 55.4 & 65.6 & 9.52\\
	& RotNet~\cite{gidaris2018unsupervised}
	& 71.9 & 94.5 & 78.4 & 70.0 & 77.2 & 86.6 & 81.6 & 93.7 & 90.7 & 88.8 & 83.3 & 8.82\\
	& GeoTrans~\cite{golan2018deep} 
	& 74.7 & \textbf{95.7} & 78.1 & 72.4 & 87.8 & \textbf{87.8} & 83.4 & \textbf{95.5} & \textbf{93.3} & \textbf{91.3} & 86.0 & 8.52\\
	\cmidrule(lr){2-2} \cmidrule(lr){3-3} \cmidrule(lr){4-4} \cmidrule(lr){5-5} \cmidrule(lr){6-6} \cmidrule(lr){7-7} \cmidrule(lr){8-8} \cmidrule(lr){9-9} \cmidrule(lr){10-10} \cmidrule(lr){11-11} \cmidrule(lr){12-12} \cmidrule(lr){13-13} \cmidrule(lr){14-14}
	& AE & 57.1 & 54.9 & 59.9 & 62.3 & 63.9 & 57.0 & 68.1 & 53.8 & 64.4 & 48.6 & 59.0 & 5.84\\
	& OURS & 78.5 & 89.8 & \textbf{86.1} & \textbf{77.4} & \textbf{90.5} & 84.5 & \textbf{89.2} & 92.9 & 92.0 & 85.5 & \textbf{86.6} & \textbf{5.35}\\
	\cmidrule(lr){1-14}
	    & GANomaly~\cite{Akcay2018} 
		& 58.9 & 57.5 & 55.7 & 57.9 & 47.9 & 61.2 & 56.8 & 58.2 & 49.7 & 48.8 & 55.3 & 4.46\\
		& Colorization~\cite{zhang2016colorful}
		& 67.6 & 62.9 & 56.8 & 62.2 & 64.7 & 68.5 & 62.2 & 63.7 & 66.5 & 71.9 & 64.7 & \textbf{4.18}\\
		ImageNet & RotNet~\cite{gidaris2018unsupervised} & 70.0 & 84.1 & 66.5 & \textbf{82.3} & \textbf{70.3} & 79.8 & 80.3 & 75.1 & 72.4 & 82.0 & 76.2 & 6.28\\
		&
		GeoTrans~\cite{golan2018deep} 
		& \textbf{72.9} & 61.0 & 66.8 & 82.0 & 56.7 & 70.1 & 68.5 & \textbf{77.2} & 62.8 & 83.6 & 70.1 & 8.43\\
		\cmidrule(lr){2-2} \cmidrule(lr){3-3} \cmidrule(lr){4-4} \cmidrule(lr){5-5} \cmidrule(lr){6-6} \cmidrule(lr){7-7} \cmidrule(lr){8-8} \cmidrule(lr){9-9} \cmidrule(lr){10-10} \cmidrule(lr){11-11} \cmidrule(lr){12-12} \cmidrule(lr){13-13} \cmidrule(lr){14-14}
		& AE & 57.1 & 51.3 & 47.7 & 57.4 & 43.8 & 54.9 & 54.6 & 51.3 & 48.3 & 41.5 & 50.8 & 5.16 \\
		& OURS & 71.9 & \textbf{85.8} & \textbf{70.7} & 78.8 & 69.5 & \textbf{83.3} & \textbf{80.6} & 72.4 & \textbf{74.9} & \textbf{84.3} & \textbf{77.2} & 5.77\\
		\cmidrule(lr){1-14}
	\end{tabular}
	\end{minipage}
	
	\begin{minipage}[t]{0.92\textwidth}
	\footnotesize
	\begin{tabular}{cx{2.0cm}x{0.7cm}x{0.7cm}x{0.7cm}x{0.7cm}x{0.7cm}x{0.7cm}x{0.7cm}x{0.7cm}x{0.7cm}x{0.7cm}x{0.8cm}}
		
		Dataset & Method & 0 & 1 & 2 & 3 & 4 & 5 & 6 & 7 & 8 & 9 & 10\\
		\cmidrule(lr){1-1} \cmidrule(lr){2-2} \cmidrule(lr){3-3} \cmidrule(lr){4-4} \cmidrule(lr){5-5} \cmidrule(lr){6-6} \cmidrule(lr){7-7} \cmidrule(lr){8-8} \cmidrule(lr){9-9} \cmidrule(lr){10-10} \cmidrule(lr){11-11} \cmidrule(lr){12-12} \cmidrule(lr){13-13}
		& DAGMM~\cite{zhai2016deep} 
		& 43.4 & 49.5 & 66.1 & 52.6 & 56.9 & 52.4 & 55.0 & 52.8 & 53.2 & 42.5 & 52.7\\
		& DSEBM~\cite{zong2018deep} 
		& 64.0 & 47.9 & 53.7 & 48.4 & 59.7 & 46.6 & 51.7 & 54.8 & 66.7 & 71.2 & 78.3 \\
		& ALOCC~\cite{Sabokrou2018Adversarially}
		& 52.7 & 56.6 & 61.2 & 61.1 & 66.7 & 50.6 & 63.9 & 66.2 & 50.9 & 73.4 & 71.1 \\
		& ADGAN~\cite{deecke2018image} 
		& 63.1 & 54.9 & 41.3 & 50.0 & 40.6 & 42.8 & 51.1 & 55.4 & 59.2 & 62.7 & 79.8 \\
		& GANomaly~\cite{Akcay2018} 
		& 57.9 & 51.9 & 36.0 & 46.5 & 46.6 & 42.9 & 53.7 & 59.4 & 63.7 & 68.0 & 75.6\\
		& GeoTrans~\cite{golan2018deep} 
		& 74.7 & 68.5 & \textbf{74.0} & \textbf{81.0} & \textbf{78.4} & 59.1 & 81.8 & 65.0 & \textbf{85.5} & \textbf{90.6} & \textbf{87.6}\\
		\cmidrule(lr){2-2} \cmidrule(lr){3-3} \cmidrule(lr){4-4} \cmidrule(lr){5-5} \cmidrule(lr){6-6} \cmidrule(lr){7-7} \cmidrule(lr){8-8} \cmidrule(lr){9-9} \cmidrule(lr){10-10} \cmidrule(lr){11-11} \cmidrule(lr){12-12} \cmidrule(lr){13-13} 
		& AE & 66.7 & 55.4 & 41.4 & 49.2 & 44.9 & 40.6 & 50.2 & 48.1 & 66.1 & 63.0 & 52.7 \\
		CIFAR-& OURS & \textbf{77.5} & \textbf{70.0} & 62.4 & 76.2 & 77.7 & \textbf{64.0} & \textbf{86.9} & \textbf{65.6} & 82.7 & 90.2 & 85.9 \\
		\cmidrule(lr){2-13}
		100 & Method & 11 & 12 & 13 & 14 & 15 & 16 & 17 & 18 & 19 & \textbf{avg} & SD\\
		\cmidrule(lr){2-2} \cmidrule(lr){3-3} \cmidrule(lr){4-4} \cmidrule(lr){5-5} \cmidrule(lr){6-6} \cmidrule(lr){7-7} \cmidrule(lr){8-8} \cmidrule(lr){9-9} \cmidrule(lr){10-10} \cmidrule(lr){11-11} \cmidrule(lr){12-12} \cmidrule(lr){13-13} 
		& DAGMM~\cite{zhai2016deep}  
		& 46.4 & 42.7 & 45.4 & 57.2 & 48.8 & 54.4 & 36.4 & 52.4 & 50.3 & 50.5 & \textbf{6.55}\\
		& DSEBM~\cite{zong2018deep}  
		& 62.7 & 66.8 & 52.6 & 44.0 & 56.8 & 63.1 & 73.0 & 57.7 & 55.5 & 58.8 & 9.36\\
		& ALOCC~\cite{Sabokrou2018Adversarially}
		& 56.9 & 63.6 & 56.0 & 57.9 & 58.2 & 57.0 & 73.5 & 61.0 & 58.8 & 60.9 & 6.70\\
		& ADGAN~\cite{deecke2018image} 
		& 53.7 & 58.9 & 57.4 & 39.4 & 55.6 & 63.3 & 66.7 & 44.3 & 53.0 & 54.7 & 10.08\\
		& GANomaly~\cite{Akcay2018} 
		& 57.6 & 58.7 & 59.9 & 43.9 & 59.9 & 64.4 & 71.8 & 54.9 & 56.8 & 56.5 & 9.94\\
		& GeoTrans~\cite{golan2018deep} 
		& \textbf{83.9} & 83.2 & 58.0 & \textbf{92.1} & 68.3 & 73.5 & \textbf{93.8} & \textbf{90.7} & 85.0 & 78.7 & 10.76 \\
		\cmidrule(lr){2-2} \cmidrule(lr){3-3} \cmidrule(lr){4-4} \cmidrule(lr){5-5} \cmidrule(lr){6-6} \cmidrule(lr){7-7} \cmidrule(lr){8-8} \cmidrule(lr){9-9} \cmidrule(lr){10-10} \cmidrule(lr){11-11} \cmidrule(lr){12-12} \cmidrule(lr){13-13}
		& AE & 62.1 & 59.6 & 49.8 & 48.1 & 56.4 & 57.6 & 47.2 & 47.1 & 41.5 & 52.4 & 8.11\\
		& OURS & 83.5 & \textbf{84.6} & \textbf{67.6} & 84.2 & \textbf{74.1} & \textbf{80.3} & 91.0 & 85.3 & \textbf{85.4} & \textbf{78.8} & 8.82\\
		\bottomrule
	\end{tabular}
	\end{minipage}
\label{tal:AUC1}
\end{table*}

\section{Experiments}
\label{sec:experiment}
We conduct substantial experiments to validate our method. Under unsupervised anomaly detection settings, the ARNet is first evaluated on multiple commonly used benchmark datasets and the large-scale dataset ImageNet~\cite{russakovsky2015imagenet}, which is rarely looked into in previous anomaly detection studies. Next, we conduct experiments on real anomaly detection datasets to evaluate the performance in real-world environments. Then we present the respective effects of different designs (\emph{e.g.},~different types of image-level transformation and loss function design) through ablation study. The stability of our models is validated through monitoring performance fluctuation during the training process and comparing the final performance after convergence in multiple training attempts, all from random weights and with the same training configuration. Finally, the visualization analysis illustrates the efficiency of the attribute restoration framework in anomaly detection.

\subsection{Experiments on Popular Benchmarks}
\noindent\textbf{Datasets.}
In this part, our experiments involve five popular image datasets: MNIST, Fashion-MNIST, CIFAR-10, CIFAR-100 and ImageNet. For all datasets, the training and test partitions remain as default. In addition, pixel values of all images are normalized to $[-1, 1]$. We introduce these five datasets briefly as follows:
\renewcommand \arraystretch{0.95}
\begin{table*}[t]
	\centering
		\caption{Average area under the ROC curve (AUROC) in \% of anomaly detection methods on MVTec AD~\cite{bergmann2019mvtec} dataset. The best performing method in each experiment is in bold.}
	\footnotesize
	\begin{tabular}{cx{0.54cm}x{0.54cm}x{0.54cm}x{0.54cm}x{0.54cm}x{0.54cm}x{0.54cm}x{0.54cm}x{0.54cm}x{0.54cm}x{0.54cm}x{0.54cm}x{0.54cm}x{0.54cm}x{0.54cm}x{0.54cm}}
		\toprule
		 Method & 0 & 1 & 2 & 3 & 4 & 5 & 6 & 7 & 8 & 9 & 10 & 11 & 12 & 13 & 14 & avg\\
		\cmidrule(lr){1-1} \cmidrule(lr){2-2} \cmidrule(lr){3-3} \cmidrule(lr){4-4} \cmidrule(lr){5-5} \cmidrule(lr){6-6} \cmidrule(lr){7-7} \cmidrule(lr){8-8} \cmidrule(lr){9-9} \cmidrule(lr){10-10} \cmidrule(lr){11-11} \cmidrule(lr){12-12} \cmidrule(lr){13-13} \cmidrule(lr){14-14} \cmidrule(lr){15-15} \cmidrule(lr){16-16} \cmidrule(lr){17-17} 
		GeoTrans~\cite{golan2018deep} & 74.4 & 67.0 & 61.9 & 84.1 & 63.0 & 41.7 & \textbf{86.9} & 82.0 & 78.3 & 43.7 & 35.9 & \textbf{81.3} & 50.0 & 97.2 & 61.1 & 67.2\\
		GANomaly~\cite{Akcay2018} & 89.2 & \textbf{73.2} & 70.8 & 84.2 & 74.3 & \textbf{79.4} & 79.2 & 74.5 & 75.7 & 69.9 & 78.5 & 70.0 & 74.6 & 65.3 & 83.4 & 76.2\\
		AE & 65.4 & 61.9 & 82.5 & 79.9 & 77.3 & 73.8 & 64.6 & 86.8 & 63.9 & 64.1 & 73.1 & 63.7 & 99.9 & 76.9 & \textbf{97.0} & 75.4\\
		OURS & \textbf{94.1} & 68.1 & \textbf{88.3} & \textbf{86.2} & \textbf{78.6} & 73.5 & 84.3 & \textbf{87.6} & \textbf{83.2} & \textbf{70.6} & \textbf{85.5} & 66.7 & \textbf{100} & \textbf{100} & 92.3 & \textbf{83.9}\\
		\bottomrule
		\end{tabular}
	\label{tal:MVTec}
\end{table*}
\begin{itemize}
    \item
    MNIST~\cite{lecun1998mnist}: consists of 70,000 $28\times28$ handwritten grayscale digit images.
    \item
    Fashion-MNIST~\cite{xiao2017fashion}: a relatively new dataset comprising $28\times28$ grayscale images of 70,000 fashion products from 10 categories, with 7,000 images per category. 
    \item CIFAR-10~\cite{krizhevsky2009learning}: consists of 60,000 $32\times32$ RGB images of 10 classes, with 6,000 images for per class. There are 50,000 training images and 10,000 test images, divided in a uniform proportion across all classes. 
    \item
    CIFAR-100~\cite{krizhevsky2009learning}: consists of 100 classes, each of which contains 600 RGB images. The 100 classes in the CIFAR-100 are grouped into 20 ``superclasses'' to make the experiment more concise and data volume of each selected ``normal class'' larger. 
    \item
    ImageNet~\cite{russakovsky2015imagenet}: We group the data from the ILSVRC 2012 classification dataset~\cite{russakovsky2015imagenet} into 10 superclasses by merging similar category labels using Latent Dirichlet Allocation (LDA)~\cite{blei2003latent}, a natural language processing method (see appendix for more details). We note that few anomaly detection research has been conducted on ImageNet since its images have higher resolution and more complex background.
\end{itemize}

\noindent\textbf{Model configuration.} The detailed structure of the model we used can be found in the appendix. We follow the settings in~\cite{unet, akccay2019skip,isola2017image} and add skip-connections between some layers in encoder and corresponding decoder layers to facilitate the backpropagation of the gradient in an attempt to improve the performance of image restoration. We use stochastic gradient descent (SGD)~\cite{bottou2010large} optimizer with default hyperparameters in Pytorch. ARNet is trained using a batch size of 32 for $500/T$ epochs, where $T$ means the number of transformations we used. The learning rate is initially set to 0.1, and is divided by 2 every $50/T$ epoch. 

In our experiments, we use a attribute erasing operation set which contains two cascade operations:
\begin{itemize}
    \item {\bf Graying:} This operation averages each pixel value along the channel dimension of images.
    \item {\bf Random rotation:} This operation rotates $x$ anticlockwise by angle $\alpha$ around the center of each image channel. The rotation angle $\alpha$ is randomly selected from a set $\{0^{\circ}, 90^{\circ},180^{\circ},270^{\circ}\}$.
\end{itemize}

The graying operation erases color information, and the random rotation operation erases objects' orientation. Both of them meet the assumptions we introduced in Section~\ref{method}. 

\noindent\textbf{Evaluation protocols.} 
For a dataset with $C$ classes, a batch of $C$ experiments are performed respectively with each of the $C$ classes set as ``normal''. We then evaluate performance on an independent test set, which contains samples from all classes, including normal and anomalous data. As all classes have equal volumes of samples in our selected datasets, the overall number proportion of normal and anomalous samples is simply $1:C-1$. 
We quantify the model performance using the area under the Receiver Operating Characteristic (ROC) curve metric (AUROC). It is commonly adopted as performance measurement in anomaly detection tasks and eliminates the subjective decision of threshold value to divide the ``normal'' samples from the anomalous ones. 

\noindent \textbf{Comparison with State-of-the-art Methods}.
Table~\ref{tal:AUC1} provides results on MNIST, Fashion-MNIST, CIFAR-10, ImageNet and CIFAR-100 in detail. Some popular methods are involved in comparison: VAE~\cite{kingma2013auto}, DAGMM~\cite{zhai2016deep}, DSEBM~\cite{zong2018deep},
ALOCC~\cite{Sabokrou2018Adversarially}, AnoGAN~\cite{schlegl2017unsupervised}, ADGAN~\cite{deecke2018image}, GANomaly~\cite{Akcay2018}, OCGAN~\cite{perera2019ocgan}, GeoTrans~\cite{golan2018deep} and our baseline backbone AE. Results of VAE, AnoGAN and ADGAN are borrowed from~\cite{deecke2018image}. 
Results of DAGMM, DSEBM and GeoTrans are borrowed from~\cite{golan2018deep}. We use the officially released source codes to fill the incomplete results reported in~\cite{Sabokrou2018Adversarially,Akcay2018} with our experimental settings. For RGB datasets, such as CIFAR-10 and CIFAR-100, we use graying and random rotation operations tandemly, together with some standard data augmentations (flipping / mirroring / shifting), which is widely used in~\cite{he2016deep, huang2017densely}. For grayscale datasets, such as MNIST and Fashion-MNIST, we only use random rotation transformation, without any data augmentation. 

\begin{figure}[tbp]
\centering
\includegraphics[width=8.3cm]{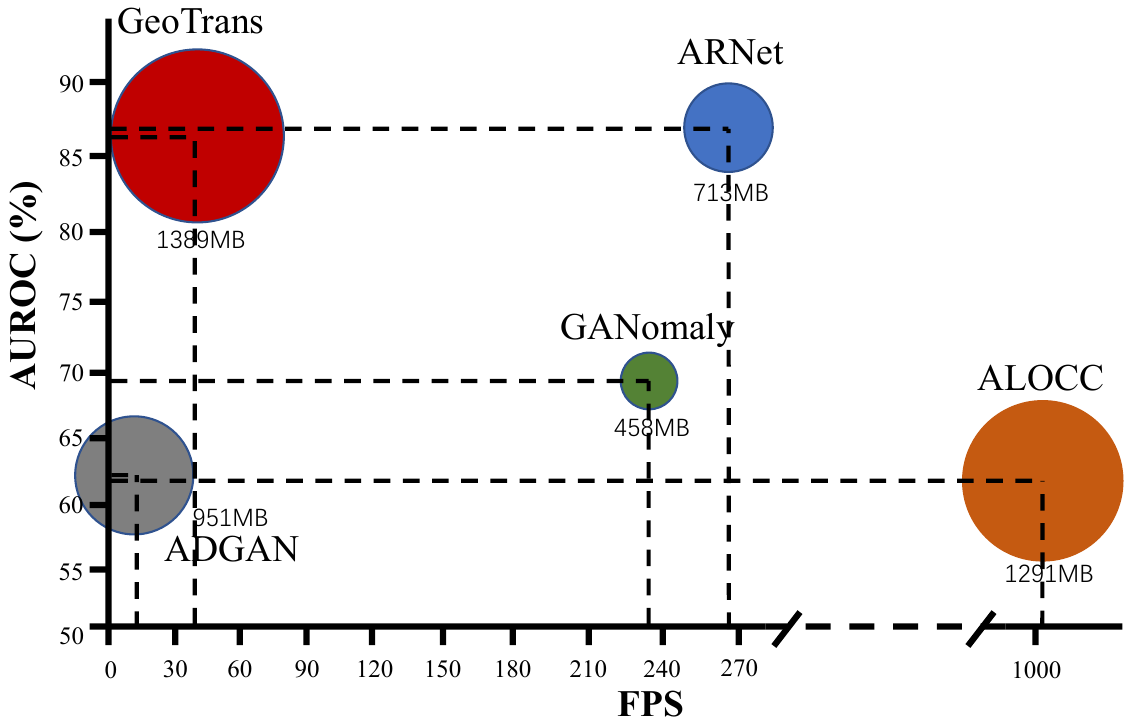}
\caption{Comparison of frames per second (FPS) (horizontal coordinates), GPU memory usages (circular sizes) and AUROC for anomaly detection (vertical coordinates) of various methods testing on CIFAR-10. ARNet takes up a relatively small GPU memory, and its FPS is relatively higher.}
\label{img:speed}
\end{figure}

\renewcommand \arraystretch{0.95}
\begin{table*}[t]
	\centering
	\caption{Average area under the ROC curve (AUROC) in \% of anomaly detection methods for \textbf{different components} on CIFAR-10. ``S'', ``G'' and ``R'' represent scaling, graying and random rotation operations. The best performing method in each experiment is in bold.}
	\small
	\begin{tabular}{cx{0.6cm}x{0.6cm}x{0.6cm}x{0.6cm}x{0.6cm}x{0.6cm}x{0.6cm}x{0.6cm}x{0.6cm}x{0.6cm}x{0.6cm}}
		\toprule
		Method & 0 & 1 & 2 & 3 & 4 & 5 & 6 & 7 & 8 & 9 & avg\\
		\cmidrule(lr){1-1} \cmidrule(lr){2-2} \cmidrule(lr){3-3} \cmidrule(lr){4-4} \cmidrule(lr){5-5} \cmidrule(lr){6-6} \cmidrule(lr){7-7} \cmidrule(lr){8-8} \cmidrule(lr){9-9} \cmidrule(lr){10-10} \cmidrule(lr){11-11} \cmidrule(lr){12-12} 
		AE (reconstruction) & 57.1 & 54.9 & 59.9 & 62.3 & 63.9 & 57.0 & 68.1 & 53.8 & 64.4 & 48.6 & 59.3\\
		ARNet+S & 72.8 & 41.8 & 66.4 & 57.5 & 71.0 & 62.8 & 68.4 & 48.5 & 56.8 & 31.9 & 57.8\\
		ARNet+G & 67.4 & 60.9 & 60.5 & 67.1 & 67.0 & 65.5 & 70.7 & 69.3 & 69.7 & 61.0 & 65.6\\
		ARNet+R & 76.1 & 80.0 & 83.6 & 77.1 & 89.2 & 83.0 & 82.6 & 85.0 & 90.0 & 75.9 & 82.2\\
		ARNet+G+R & \textbf{78.5} & \textbf{89.8} & \textbf{86.1} & \textbf{77.4} & \textbf{90.5} & \textbf{84.5} & \textbf{89.2} & \textbf{92.9} & \textbf{92.0} & \textbf{85.5} & \textbf{86.6}\\
		\bottomrule
		\end{tabular}
	\label{tal:AUC_ablation}
\end{table*}

\renewcommand \arraystretch{0.95}
\begin{table*}[t]
    \newcommand{\tabincell}[2]{\begin{tabular}{@{}#1@{}}#2\end{tabular}}
	\centering
	\caption{Average area under the ROC curve (AUROC) in \% of anomaly detection methods for \textbf{different losses} on CIFAR-10. ``$\ell_1$'' means $\ell_1$ loss and ``$\ell_2$'' means $\ell_2$ loss. For example, $\ell_2\rightarrow \ell_1$ means using $\ell_2$ loss as training loss to train autoencoders and using $\ell_1$ loss to calculate restoration error when testing. The best performing method in each experiment is in bold.}
	\small
	\begin{tabular}{cx{1.5cm}x{1.5cm}x{1.5cm}x{2.5cm}x{0.4cm}x{0.4cm}x{0.4cm}x{0.4cm}x{0.4cm}x{0.4cm}x{0.4cm}x{0.4cm}}
		\toprule
		$c_i$ & $\ell_1\rightarrow \ell_1$ & $\ell_1\rightarrow \ell_2$ & $\ell_2\rightarrow \ell_2$ & $\ell_2\rightarrow \ell_1$(OURS)\\
		\cmidrule(lr){1-1} \cmidrule(lr){2-2} \cmidrule(lr){3-3} \cmidrule(lr){4-4} \cmidrule(lr){5-5}
		0 & 74.2 & 74.1 & 77.8 & \textbf{78.5}\\
		1 & 82.0 & 80.7 & 86.8 & \textbf{89.8}\\
		2 & 82.6 & 81.9 & 85.2 & \textbf{86.1}\\
		3 & 77.2 & 77.1 & 76.0 & \textbf{77.4}\\
        \bottomrule
	\end{tabular}
	\label{tal:AUC_loss}
\end{table*}

On all involved datasets, experiment results present that the average AUROC of ARNet outperforms all other methods to different extents. For each individual image class, we also obtain competitive performances, showing effectiveness for anomaly detection. To further validate the effectiveness of our method, we conduct experiments on a subset of the ILSVRC 2012 classification dataset~\cite{russakovsky2015imagenet}. Table~\ref{tal:AUC1} also shows the performance of GANomaly, GeoTrans, baseline AE and our method on ImageNet. As can be seen, our method significantly outperforms the other three methods and maintains performance stability on more difficult datasets. We further compared to two self-supervised approaches, Colorization~\cite{zhang2016colorful} and RotNet~\cite{gidaris2018unsupervised}, for anomaly detection. For each method, we train the model with the dataset containing only normal data. When testing, we utilize the original loss of each method as the anomaly score, and the sample corresponding to a large loss is considered as anomalous. As shown in Table~\ref{tal:AUC1}, ARNet outperforms the two self-supervised methods.

\noindent \textbf{Computational Cost}. We investigate the computational efficiency and the GPU memory cost. For all methods, we test for 10 times on CIFAR-10 (total 10,000 images) with NVIDIA GTX 1080Ti and record the average FPS and the GPU memory costs, the results are shown in Figure~\ref{img:speed}. ALOCC~\cite{Sabokrou2018Adversarially} has an advantage in the highest computational efficiency (1049fps) but takes up relatively large GPU memory (1291MB). GeoTrans~\cite{golan2018deep} takes up more GPU memory (1389MB) and suffering from a slow computational efficiency (35fps). ARNet reaches 270fps ($5\times$ faster than GeoTrans) and takes only 713MB of GPU memory. Though during the testing phase, ARNet needs to traverse all the transformations functions, ARNet still reaches a considerable efficiency thanks to its light network structure. With the best performance in AUROC, ARNet takes up a relatively small GPU memory with high computational efficiency.

\subsection{Experiments on Real-world Anomaly Detection}

Previous works~\cite{golan2018deep,deecke2018image} experiment on multi-class classification datasets since there is a lack of comprehensive real-world datasets available for anomaly detection. By defining anomalous events as occurrences of different object classes and splitting the datasets based on unsupervised settings, the multi-class datasets can be used for anomaly detection experiments. However, the real anomalous data does not necessarily meet the above settings, \emph{e.g.},~damaged objects. In this section, we experiment on the most recent real-world anomaly detection benchmark dataset MVTec AD~\cite{bergmann2019mvtec}.

\renewcommand \arraystretch{0.95}
\begin{table*}[!htb]
	\centering
	\caption{Average area under the ROC curve (AUROC) in \% of anomaly detection methods on MNIST for ten runs in which digit number ``1'' is taken as normal data. Our stability is much higher than GeoTrans.}
	\small
	\begin{tabular}{cx{0.8cm}x{0.8cm}x{0.8cm}x{0.8cm}x{0.8cm}x{0.8cm}x{0.8cm}x{0.8cm}x{0.8cm}x{0.8cm}x{0.8cm}x{0.7cm}}
	\toprule
	Methods & \#1 & \#2 & \#3 & \#4 & \#5 & \#6 & \#7 & \#8 & \#9 & \#10 & avg & SD\\
	\cmidrule(lr){1-1} \cmidrule(lr){2-2} \cmidrule(lr){3-3} \cmidrule(lr){4-4} \cmidrule(lr){5-5} \cmidrule(lr){6-6} \cmidrule(lr){7-7} \cmidrule(lr){8-8} \cmidrule(lr){9-9} \cmidrule(lr){10-10} \cmidrule(lr){11-11} \cmidrule(lr){12-12} \cmidrule(lr){13-13}
	GeoTrans
	~\cite{golan2018deep} 
	& 91.55 & 72.38 & 81.26 & 82.94 & 87.04 & 87.95 & 87.24 & 81.77 & 85.51 & 85.68 & 84.33 & 5.22\\
	%\hline
	OURS & 99.93 & 99.94 & 99.95 & 99.94 & 99.95 & 99.93 & 99.93 & 99.94 & 99.92 & 99.93 & 99.94 & 0.01\\
	\bottomrule
	\end{tabular}
	\label{tal:stable2}.
\end{table*}

\noindent\textbf{MVTec anomaly detection dataset.} MVTec Anomaly Detection (MVTec AD) dataset~\cite{bergmann2019mvtec} contains 5354 high-resolution color images of different object and texture categories. It contains normal images intended for training and images with anomalies intended for testing. The anomalies manifest themselves in the form of over 70 different types of defects such as scratches, dents, and various structural changes. In this paper, we conduct image-level anomaly detection tasks on the MVTec AD dataset to classify normal and anomalous objects.

\noindent\textbf{Comparison with state-of-the-art methods.} Table~\ref{tal:MVTec} shows that ARNet performs better than baseline AE, GANomaly and GeoTrans. The advantages of ARNet over GeoTrans are growing from ideal datasets to real-world datasets MVTec AD. We conclude that ARNet is more adaptable to complex real-world environments. 

\subsection{Ablation Study}
We study the contribution of the proposed components of ARNet independently. Table~\ref{tal:AUC_ablation} shows experimental results of ablation study on CIFAR-10. It shows that both graying and random rotation operations improve the performance significantly, especially the random rotation operation. Table~\ref{tal:AUC_loss} shows the ablation study about the selection of restoration loss. It proves that using $\ell_2$ loss as training loss and using $\ell_1$ loss to calculate restoration error performs the best. Through the ablation study, we claim that the attribute erasing operations, network architecture and the loss function we used all have independent contributions to boost the model performance.

We use the image scaling to study the degradation problem of the ARNet caused by ill-selected attribute erasing operation. Downsampling of images can delete part of the image information, but it is not correlated to any object attributes. It will lead the model to learn a linear interpolation process in which both normal and anomalous data can be restored easily by inferring from neighboring pixels. Thus, this operation cannot remove any attribute which satisfies the assumptions introduced in Section~\ref{method}. In this case, the AEM module is not functional and the ARNet will degrade into a vanilla AE. We test on CIFAR-10 with a 0.5x scaling and obtain 58.8\% AUROC for ARNet, while that of AE is 59.3\%, showing that the ARNet indeed degenerates into a vanilla AE with ill-selected attribute erasing operation.

\begin{figure}[t]
\centering
\includegraphics[width=7.3cm]{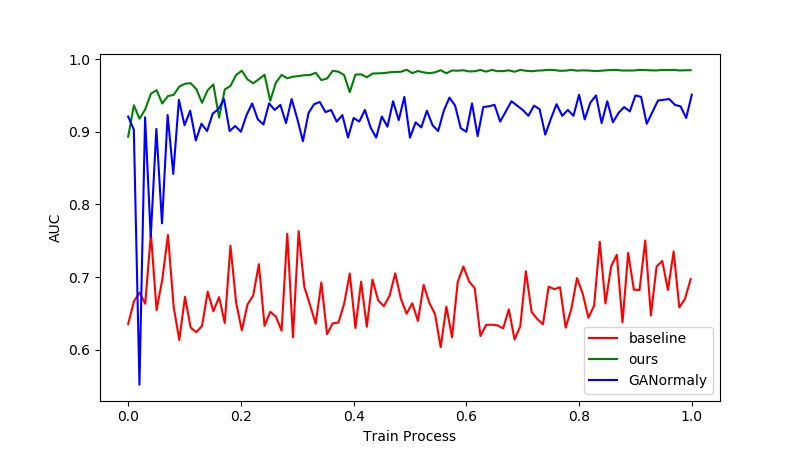}
\caption{Training process under three methods. Both logs are achieved on the MNIST dataset. It shows the case when the digit ``7'' is the normal class. We attach complete logs for Fashion-MNIST and MNIST datasets in the appendix.}
\label{fig:stable1}
\end{figure}

\begin{figure}[t]
\centering
\includegraphics[width=7.3cm]{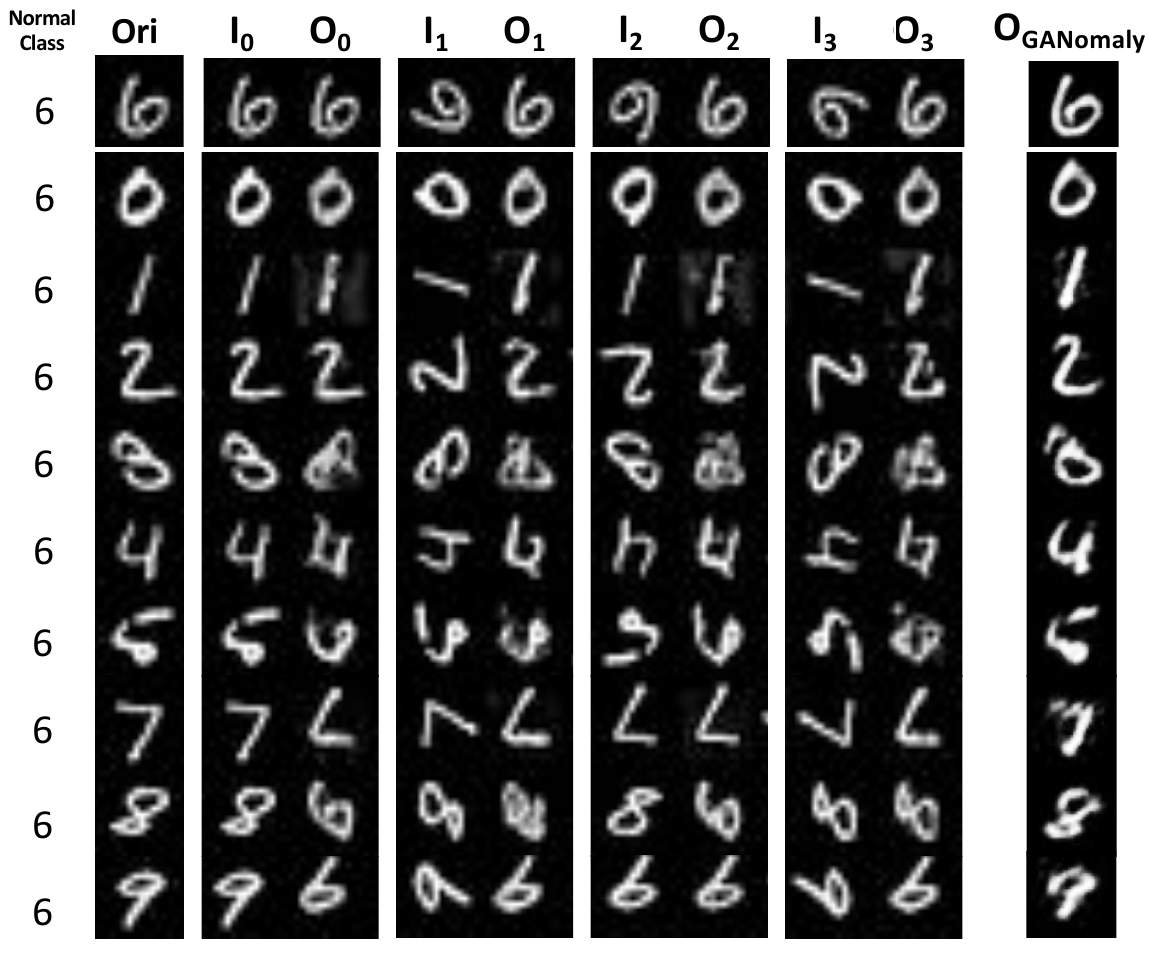}
\caption{Visualization analysis comparing with GANomaly on MNIST. ``Ori'', ``I'' and ``O'' represent original images, inputs and outputs, respectively. Cases with outputs similar to ``Ori'' are considered normal, otherwise anomalous. All visualization results are based on the number ``6'' as normal samples.}
\label{img:color}
\end{figure}

\begin{figure}[t]
\centering
\includegraphics[width=7.7cm]{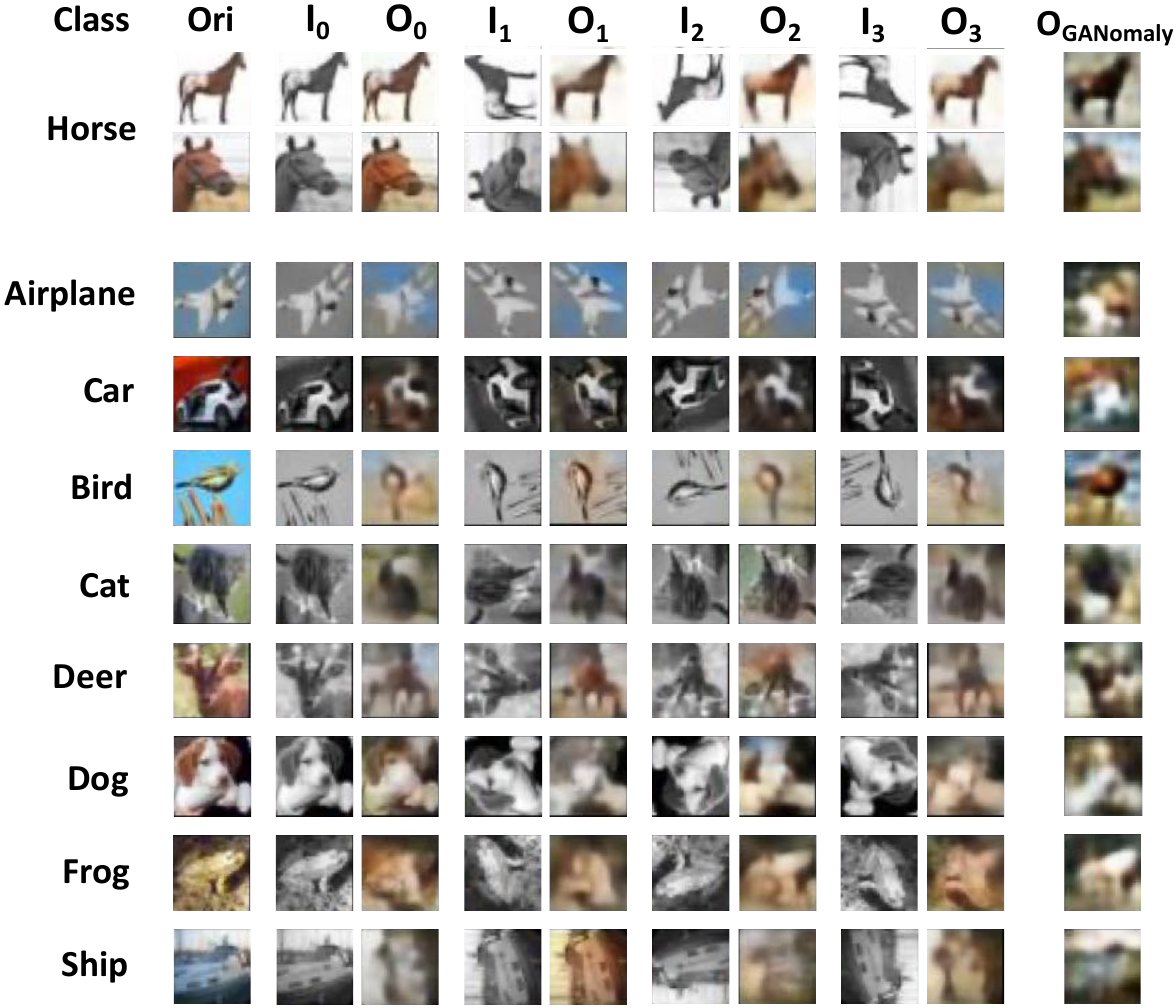}
\caption{Visualization analysis comparing with GANomaly on CIFAR-10. ``Ori'', ``I'' and ``O'' represent original images, inputs and outputs, respectively. Cases with outputs similar to ``Ori'' are considered normal, otherwise anomalous. All visualization results are based on the class ``horse'' as normal samples.}
\label{img:cifar10_vis}
\end{figure}

\subsection{Model Stability} 
Anomaly detection puts higher concerns on the stability of model performance than traditional classification tasks. It is because of the lack of anomalous data makes it impossible to do validation during training. Thus, model stability tends to be more important since without validation there is no way to select the best checkpoint for anomaly detection model in the training phases. The stability of model performance is mainly reflected in three aspects: 1) whether the model can stably reach convergence after acceptable training epochs in one training attempt; 2) whether the model can reach stable performance level in multiple independent training attempts under the same training configuration; 3) whether the model can stably achieve good performance in various datasets and training configurations. Figure~\ref{fig:stable1} shows AUC-changing during one run to reveal that our model performs more stably in the late training phase. Thus, through ARNet, a highly reliable model can be achieved through acceptable training epochs in this practically validation-unavailable task. In order to test the stability of multiple training performances, we rerun GeoTrans~\cite{golan2018deep} and our method for 10 times on MNIST. Table~\ref{tal:stable2} shows that GeoTrans suffers a larger performance fluctuation compared with our method. Table~\ref{tal:AUC1} shows that our method has the strongest stability of this type, as can be seen from its low standard deviation (SD).

\renewcommand \arraystretch{0.95}
\begin{table}[t]
	\centering
		\caption{Average area under the ROC curve (AUROC) in \% of anomaly detection methods on CIFAR-10-C dataset with \textbf{five distorted severity~\cite{hendrycks2018benchmarking}}. Each level includes all the 19 distorted categories. The last column shows the average AUROC and its anomaly detection performance degradation compared to the original CIFAR-10 test set in Figure~\ref{tal:AUC1}. The best performing method in each experiment is in bold.}
	\footnotesize
	\begin{tabular}{C{2.1cm}C{0.45cm}C{0.45cm}C{0.45cm}C{0.45cm}C{0.45cm}C{1.4cm}}
		\toprule
		 Distorted Severity & 1 & 2 & 3 & 4 & 5 & avg\\
		\cmidrule(lr){1-1} \cmidrule(lr){2-2} \cmidrule(lr){3-3} \cmidrule(lr){4-4} \cmidrule(lr){5-5} \cmidrule(lr){6-6}
		\cmidrule(lr){7-7}
		ALOCC~\cite{Sabokrou2018Adversarially} & 49.1 & 49.1 & 49.2 & 49.1 & 49.1 & 49.1 (13.1$\downarrow$)\\
		GANomaly~\cite{Akcay2018} & 48.3 & 48.4 & 48.5 & 48.7 & 48.8 & 48.8 (21.0$\downarrow$)\\
		GeoTrans~\cite{golan2018deep} & 83.5 & 81.8 & 80.6 & \textbf{79.1} & \textbf{76.3} & 80.2 (5.8$\downarrow$)\\
		ARNet (ours) & \textbf{84.3} & \textbf{82.6} & \textbf{80.9} & 78.9 & 75.5 & \textbf{80.5 (6.1$\downarrow$)}\\
		\bottomrule
		\end{tabular}
	\label{tal:cifar10c1}
\end{table}

\renewcommand \arraystretch{0.95}
\begin{table}[t]
\centering
\caption{Average area under the ROC curve (AUROC) in \% of anomaly detection methods on CIFAR-10-C dataset with \textbf{19 distorted categories~\cite{hendrycks2018benchmarking}}. For each distorted category, we conduct experiments on five levels of distorted severity, and the result is reported in average AUROC. The best performing method in each experiment is in bold.}
\label{tal:cifar10c2}
\footnotesize
\begin{tabular}{C{0.8cm}C{1.0cm}C{1.0cm}C{1.35cm}C{1.1cm}C{1.0cm}}
\toprule
& & \makecell[c]{ALOCC\\~\cite{Sabokrou2018Adversarially}} & \makecell[c]{GANomaly\\~\cite{Akcay2018}} & \makecell[c]{GeoTrans\\~\cite{golan2018deep}} & \makecell[c]{ARNet\\ (ours)}\\
\cmidrule(lr){1-1} \cmidrule(lr){2-2} \cmidrule(lr){3-3} \cmidrule(lr){4-4} \cmidrule(lr){5-5} \cmidrule(lr){6-6} 
& Gauss & 49.0 & 47.9 & 75.3 & \textbf{76.9} \\
Noise & Shot & 49.0 & 48.0 & 77.5 & \textbf{78.9}\\
& Impulse & 49.0 & 47.8 & 75.4 & \textbf{76.3}\\
& Speckle & 48.9 & 48.0 & 77.6 & \textbf{78.7}\\
\cmidrule(lr){1-1} \cmidrule(lr){2-2} \cmidrule(lr){3-3} \cmidrule(lr){4-4} \cmidrule(lr){5-5} \cmidrule(lr){6-6} 
& Gauss & 49.0 & 48.2 & \textbf{82.1} & 81.1\\
& Defocus & 49.0 & 48.2 & \textbf{83.3} & 82.8\\
Blur & Glass & 48.7 & 48.0 & \textbf{74.3} & 73.0\\
& Zoom & 48.9 & 48.2 & \textbf{83.1} & 82.1 \\
& Motion & 49.0 & 48.1 & 80.0 & \textbf{80.1} \\
\cmidrule(lr){1-1} \cmidrule(lr){2-2} \cmidrule(lr){3-3} \cmidrule(lr){4-4} \cmidrule(lr){5-5} \cmidrule(lr){6-6} 
& Snow & 49.8 & 49.7 & 79.7 & \textbf{82.4} \\
& Frost & 50.0 & 50.3 & \textbf{79.9} & 79.5\\
Weather & Fog & 49.7 & 48.1 & 82.7 & \textbf{84.3}\\
& Spatter & 49.0 & 48.0 & \textbf{82.0} & 81.5\\
& Bright & 49.7 & 49.8 & \textbf{84.5} & \textbf{84.5} \\
& Saturate & 49.0 & 50.0 & \textbf{84.4} & 83.5 \\
\cmidrule(lr){1-1} \cmidrule(lr){2-2} \cmidrule(lr){3-3} \cmidrule(lr){4-4} \cmidrule(lr){5-5} \cmidrule(lr){6-6} 
& Contrast & 48.0 & 50.1 & \textbf{78.3} & 76.4\\
Digital & Elastic & 48.8 & 48.0 & 82.0 & \textbf{82.1}\\
& Pixelate & 48.9 & 48.1 & 82.3 & \textbf{82.6} \\
& JPEG & 49.0 & 48.1 & 80.0 & \textbf{82.2}\\
\bottomrule
\end{tabular}
\end{table}

\begin{figure*}[tbph]
  \begin{minipage}[t]{0.158\textwidth}
\centering
\includegraphics[width=2.8cm]{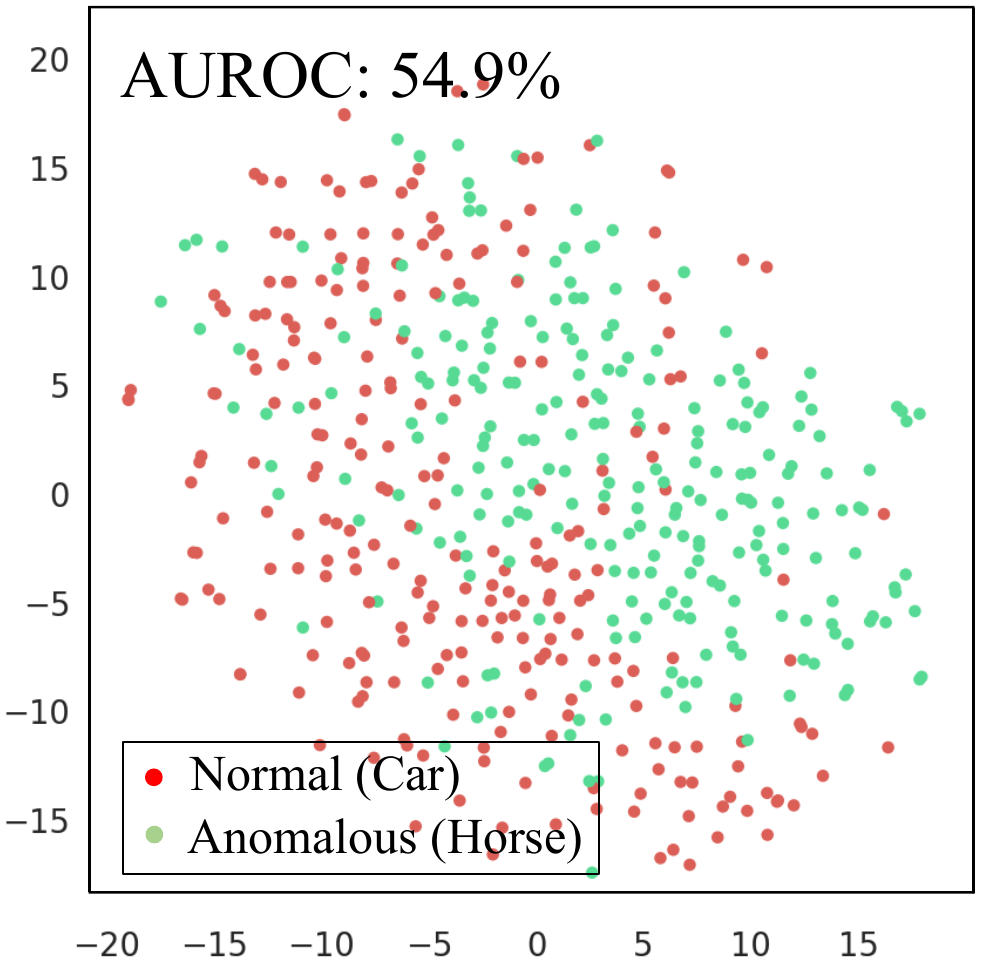}
\footnotesize
(a) 1-Autoencoder
\end{minipage}
\begin{minipage}[t]{0.158\textwidth}
\centering
\includegraphics[width=2.8cm]{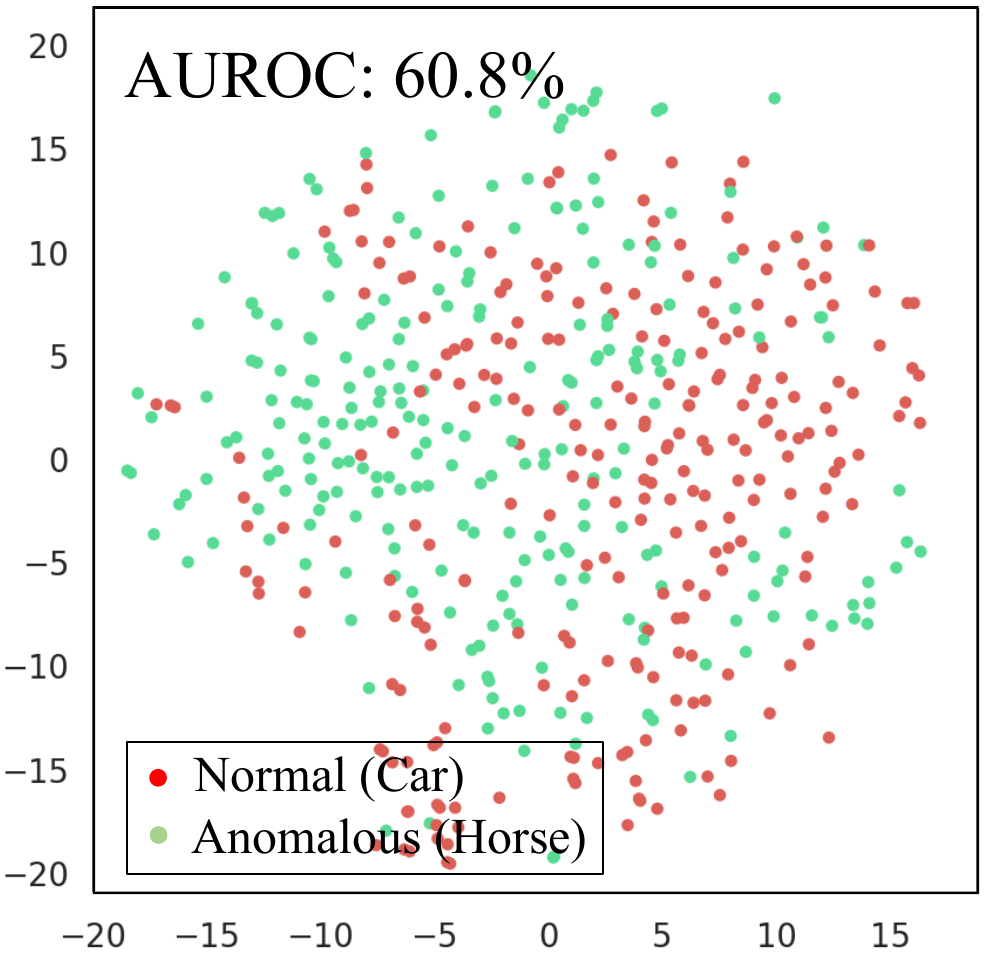}
\footnotesize
(b) 1-GANomaly
\end{minipage}
\begin{minipage}[t]{0.158\textwidth}
\centering
\includegraphics[width=2.8cm]{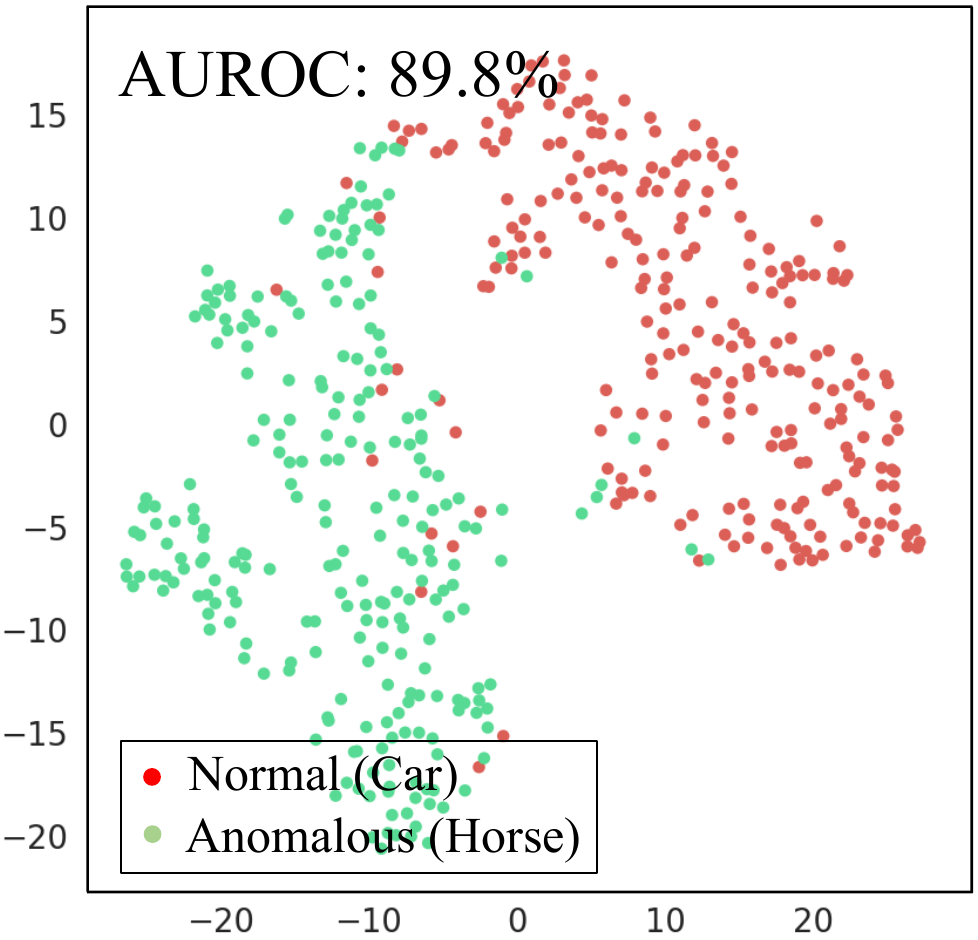}
\footnotesize
(c) 1-ARNet
\end{minipage}
\begin{minipage}[t]{0.158\textwidth}
\centering
\includegraphics[width=2.8cm]{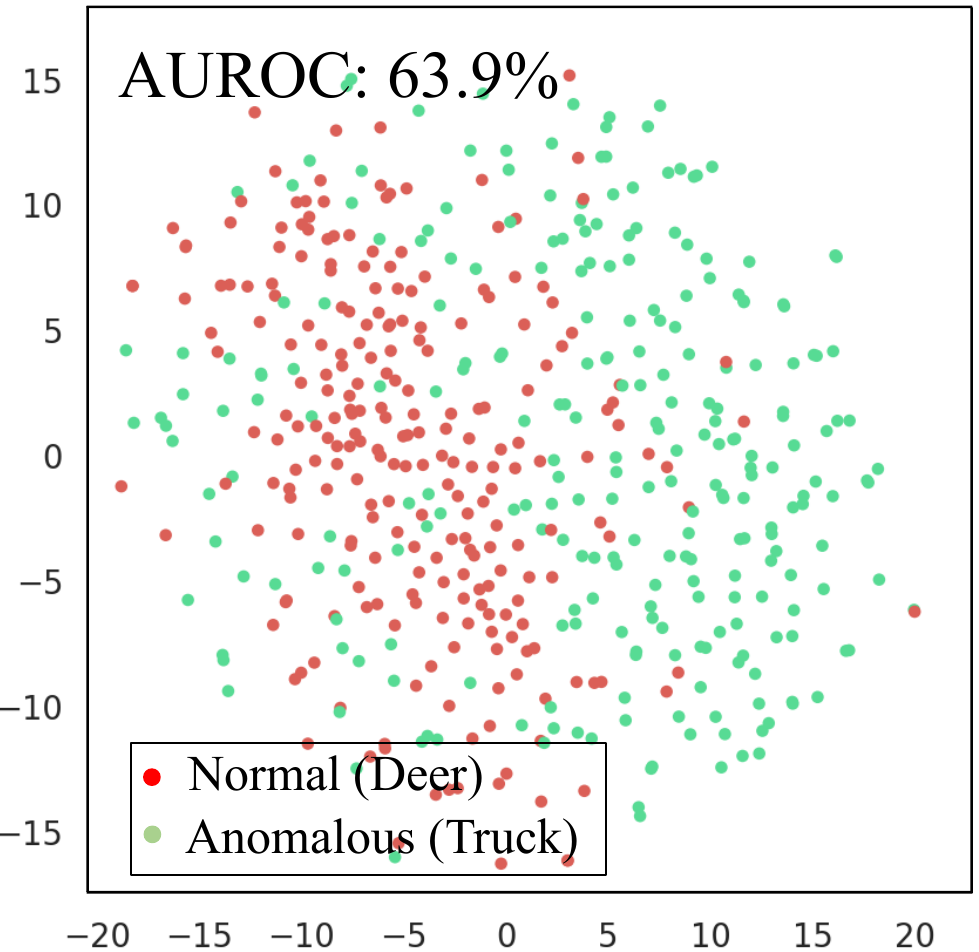}
\footnotesize
(d) 2-Autoencoder
\end{minipage}
\begin{minipage}[t]{0.158\textwidth}
\centering
\includegraphics[width=2.8cm]{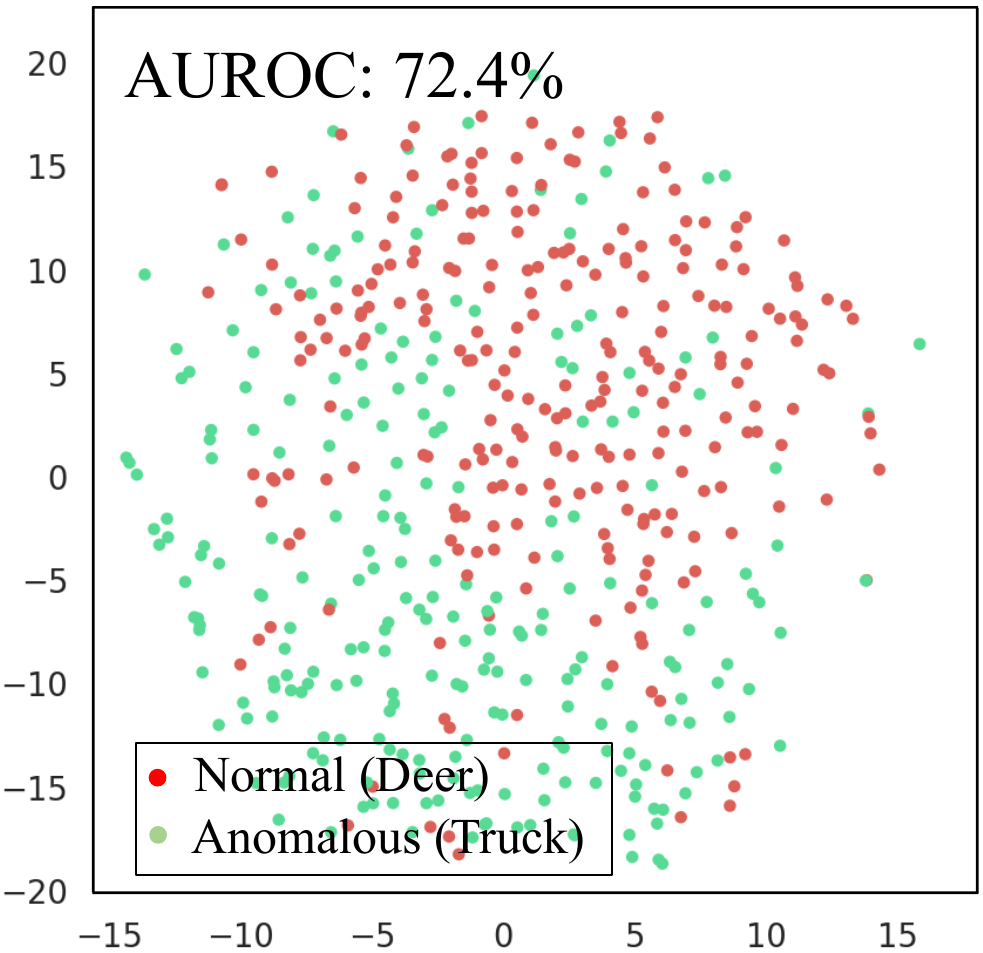}
\footnotesize
(e) 2-GANomaly
\end{minipage}
\begin{minipage}[t]{0.158\textwidth}
\centering
\includegraphics[width=2.8cm]{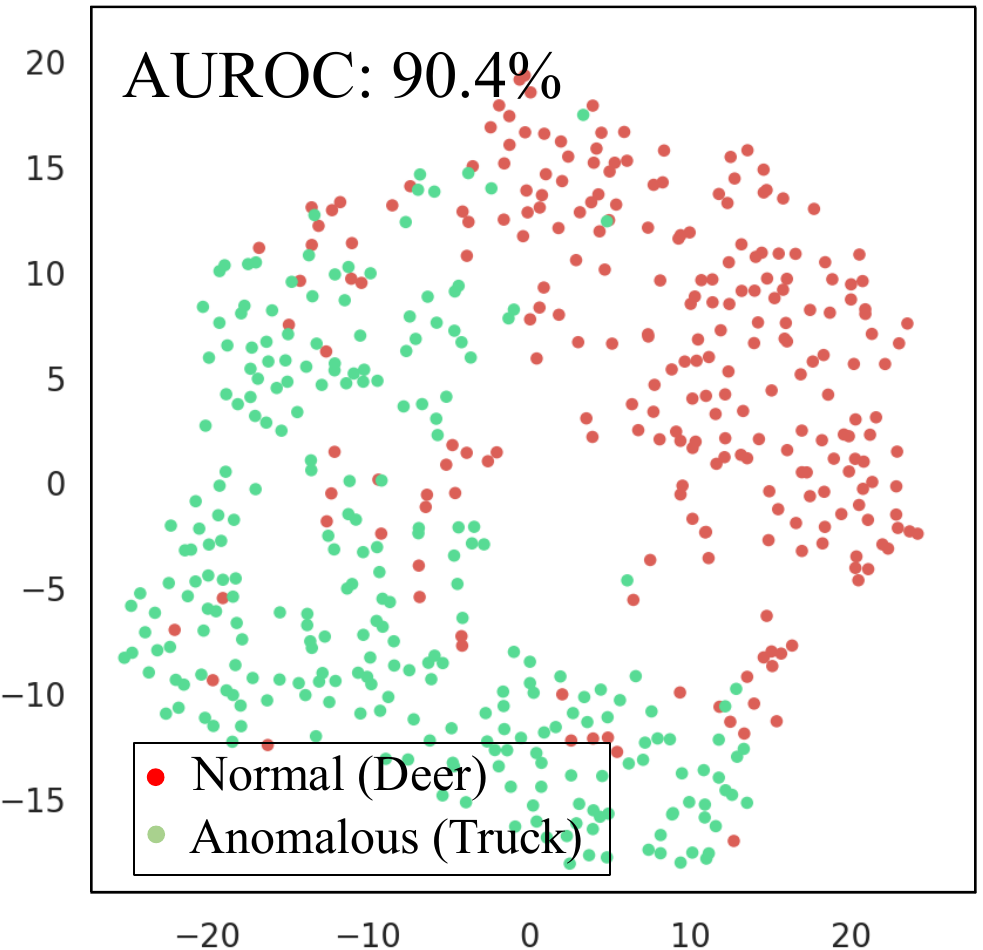}
\footnotesize
(f) 2-ARNet
\end{minipage}

\begin{minipage}[t]{0.158\textwidth}
\centering
\includegraphics[width=2.8cm]{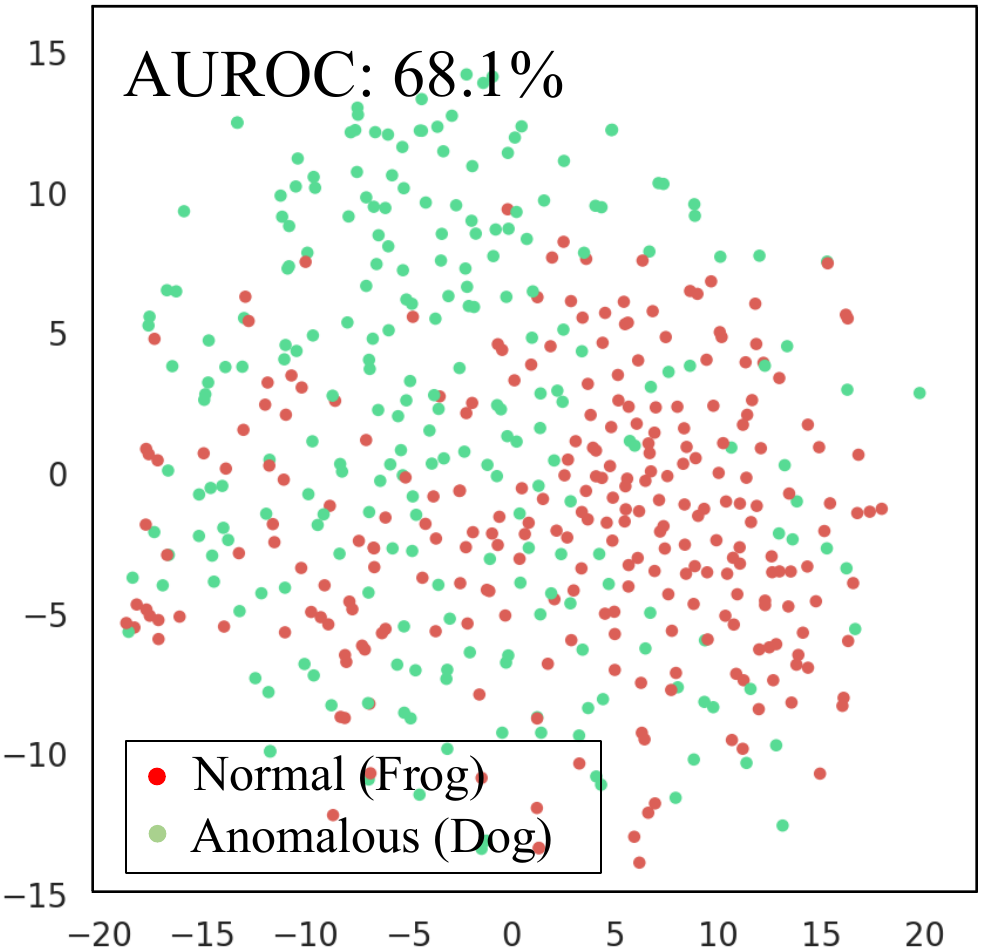}
\footnotesize
(g) 3-Autoencoder
\end{minipage}
\begin{minipage}[t]{0.158\textwidth}
\centering
\includegraphics[width=2.8cm]{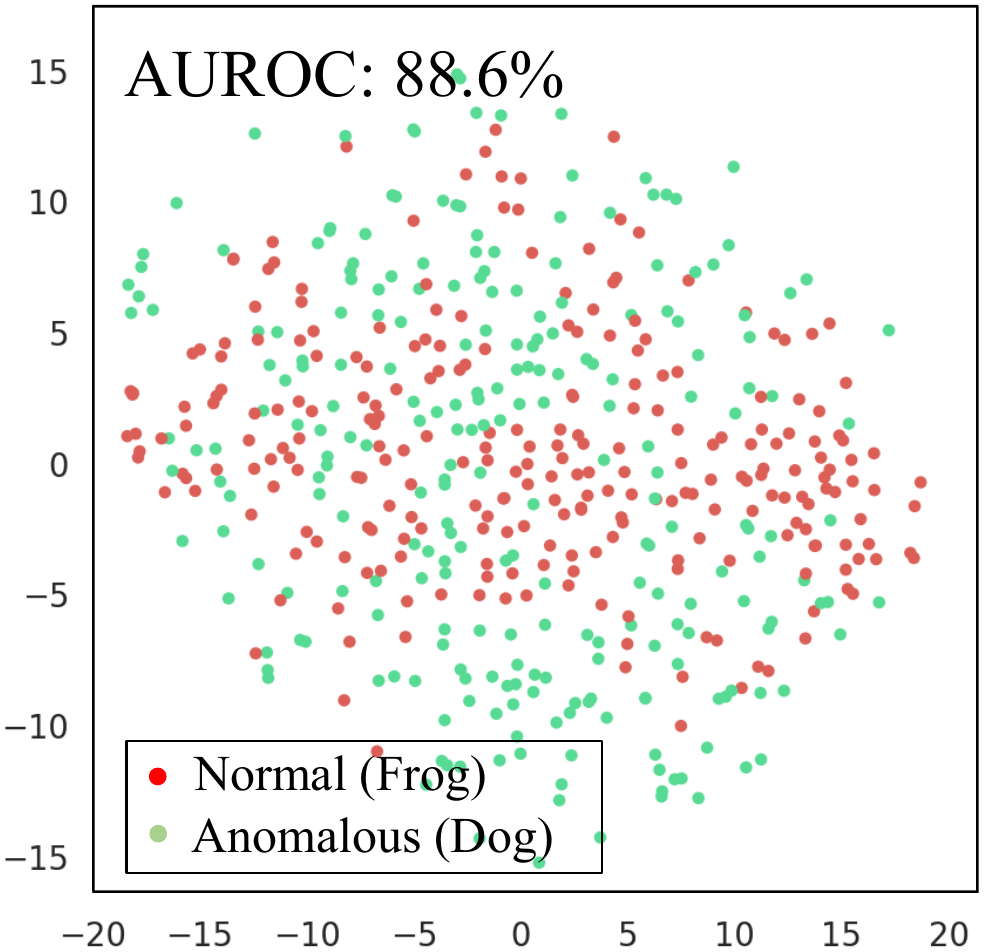}
\footnotesize
(h) 3-GANomaly
\end{minipage}
\begin{minipage}[t]{0.158\textwidth}
\centering
\includegraphics[width=2.8cm]{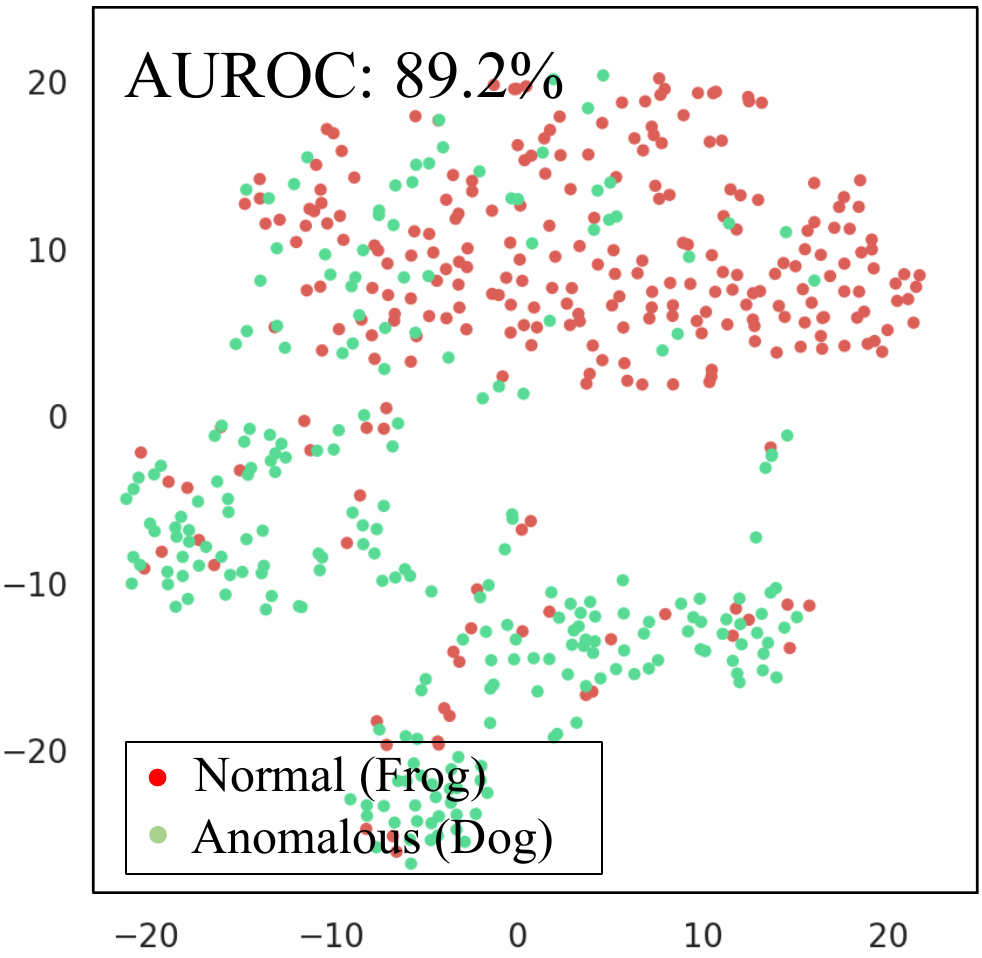}
\footnotesize
(i) 3-ARNet
\end{minipage}
\begin{minipage}[t]{0.158\textwidth}
\centering
\includegraphics[width=2.8cm]{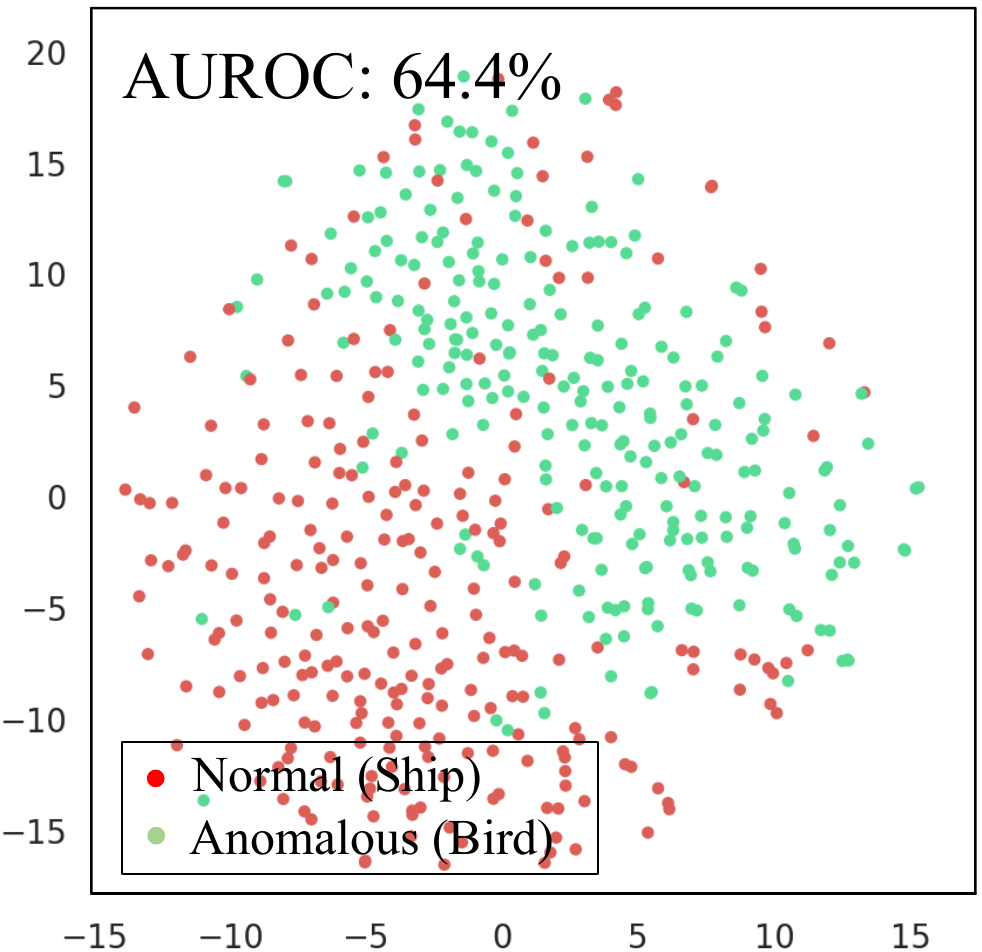}
\footnotesize
(j) 4-Autoencoder
\end{minipage}
\begin{minipage}[t]{0.158\textwidth}
\centering
\includegraphics[width=2.8cm]{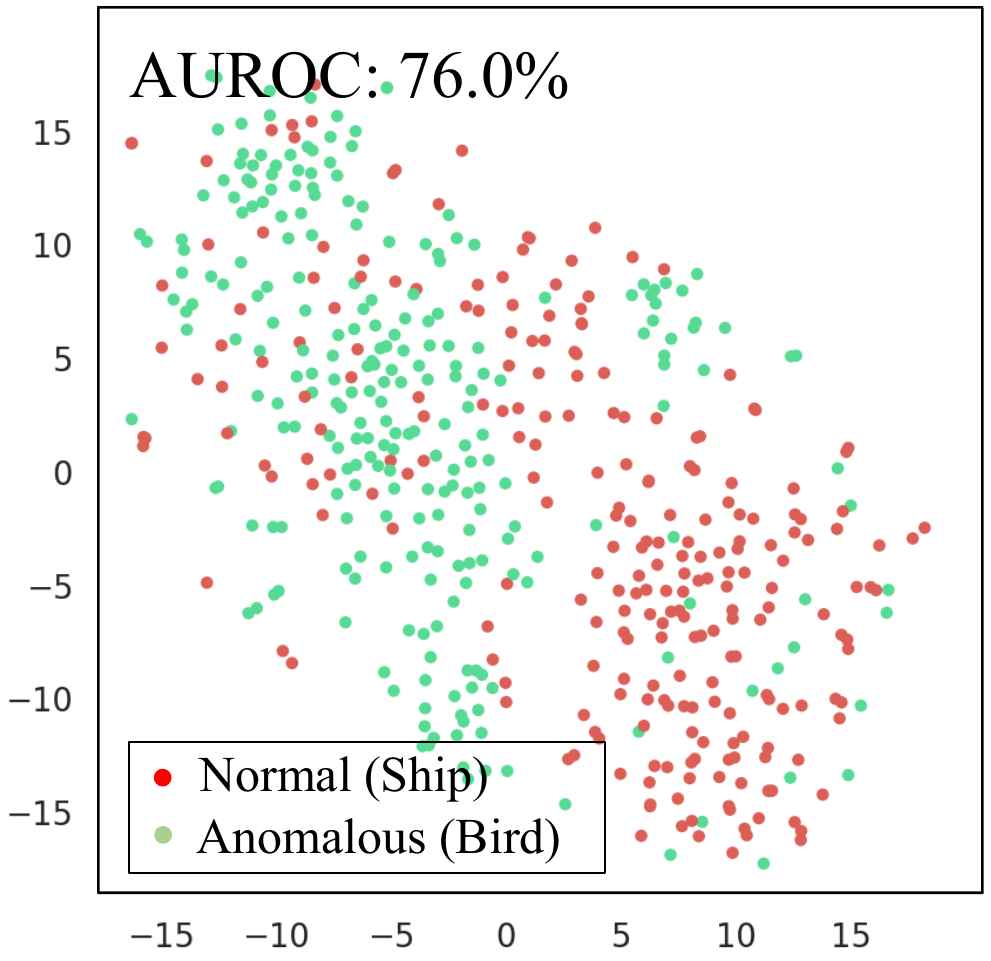}
\footnotesize
(k) 4-GANomaly
\end{minipage}
\begin{minipage}[t]{0.158\textwidth}
\centering
\includegraphics[width=2.8cm]{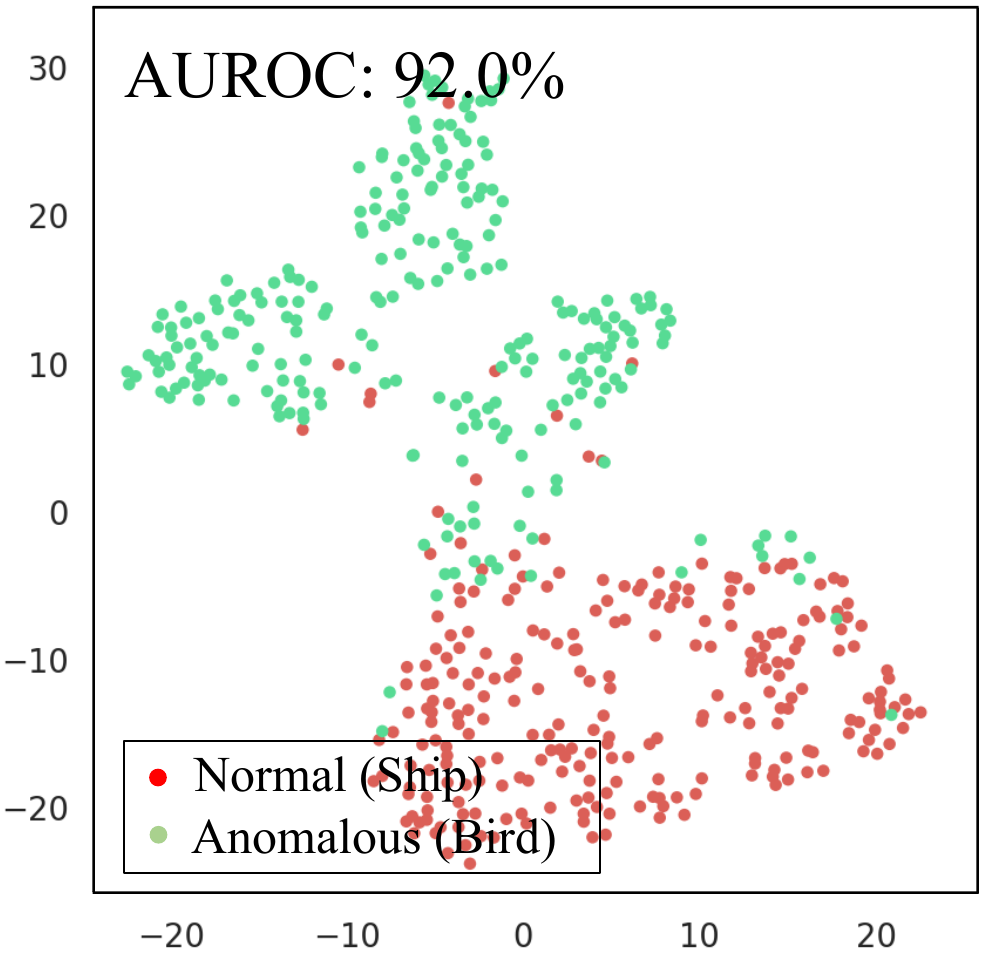}
\footnotesize
(l) 4-ARNet
\end{minipage}
\caption{T-SNE visualization of latent spaces of autoencoder, GANomaly and ARNet on CIFAR-10. The corresponding AUROCs of anomaly detection are marked in the upper left corners.}
  \label{fig:tsne_cifar}
\end{figure*}

\begin{figure}[tbp]
\centering
\includegraphics[width=7.3cm]{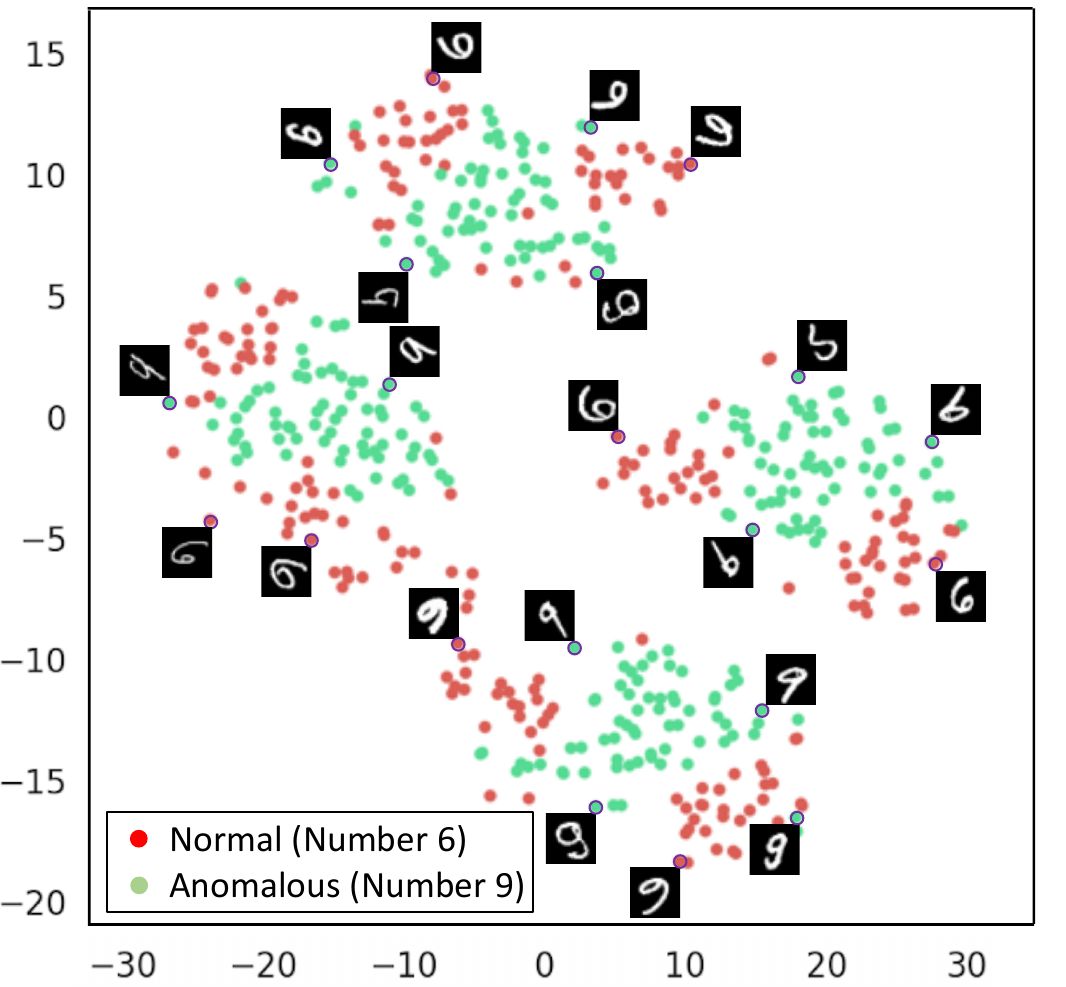}
\caption{T-SNE visualization of latent spaces of ARNet on number 6 and 9 in the handwritten dataset MNIST. Number 6 is set as the normal class.}
\label{img:tsne_mnist}
\end{figure}

\subsection{Visualization Analysis}
In order to demonstrate the effectiveness of the attribute restoration framework for anomaly detection in a simple and straightforward way, we visualize some restoration outputs from ARNet, comparing with GANomaly in Figure~\ref{img:color} and Figure~\ref{img:cifar10_vis}. 
For MNIST and CIFAR-10, all visualization results are based on the same experimental setting in which the number ``6'' and the class ``horse'' are considered normal samples respectively.

The first column ``Ori'' represents original images. ``I'' means images after Attribute Erasing Module. Note that the restoration error is calculated between outputs and original images. Cases with low restoration error with original images are considered normal, otherwise anomalous. For example, the bottom line in Figure~\ref{img:color} shows the testing results of number ``9''. Intuitively, four outputs are far different from ``Ori'' and thus recognized as anomalous. Except for the number ``6'', the other numbers get either wrong direction or ambiguous restoration outputs from ARNet. It enlarges the gap of restoration error between normal and anomalous data. However, all the outputs from GANomaly are similar to the ground truth, meaning that it is less capable to distinguish between normal and anomalous data. In addition, for the anomalous cases in Figure~\ref{img:cifar10_vis}, restoration errors even larger since the wrong colors the ARNet used for image restoration. All the outputs show that ARNet attempts to restore the input images using the orientation or color distribution of the normal classes learning from the training set.

\section{Extended Experiments and Discussion}
\subsection{Can ARNet Extract Semantic Features?}
ARNet has achieved better performance by introducing the restoration framework into anomaly detection. Here we further investigate whether ARNet can extract the semantic features.

The CIFAR-10-C~\cite{hendrycks2018benchmarking} is designed to measure a method's robustness to some common image corruptions. It is obtained by applying different corruptions to CIFAR-10. There are in total 19 diverse corruption types and each type of corruption has five levels of severity, resulting in 95 distinct corruptions. With the distortions, the models with less performance drop are more likely to be able to extract semantic features, as the low-level information in the original images has been damaged to a certain extent. We train all models with CIFAR-10 and test them on the CIFAR-10-C test set. Table~\ref{tal:cifar10c1} and Table~\ref{tal:cifar10c2} provide the comparison of ARNet with several state-of-the-art methods ALOCC~\cite{Sabokrou2018Adversarially}, GANomaly~\cite{akccay2019skip} and GeoTrans~\cite{golan2018deep}. As can be seen from Table~\ref{tal:cifar10c1}, ARNet and GeoTrans are robust to corruptions of input images and with only 6.1\% and 5.8\% performance drop on CIFAR-10-C, respectively, suggesting that the features extracting by ARNet and GeoTrans have more semantic information. On the contrary, both ALOCC and GANomaly have much more significant performance drop in this case, indicating that they rely heavily on the low-level information.

We further show that the latent representations extracted by ARNet can be used to distinguish samples of different categories through T-SNE analysis, i.e. the extracting features already contain semantic information about the different categories. We first show T-SNE visualization results of CIFAR-10. As shown in Figure~\ref{fig:tsne_cifar}, feature maps of latent space for ARNet are more discriminative than those of AE and GANomaly, suggesting that ARNet can extract more semantic features than the other two methods. Figure~\ref{img:tsne_mnist} further shows a more specific case with handwritten numbers 6 and 9 from MNIST,  where numbers 6 are set to be normal. We use the random rotation operation in this task. As can be seen from Figure~\ref{img:tsne_mnist}, T-SNE clusters the data into four categories, corresponding to the four degrees of rotations on normal data. For example, the number 6 without rotation and the number 9 rotated 180 degrees are grouped into the same category. In order to restore the handwritten number 6 to the original images with the correct orientation, the decoder needs to simply map these four categories to the one category which has the same orientation. And this simple mapping operation will cause a large image restoration error for the number 9. 

\subsection{Do the attributes easily be shared between normal and anomalous data?}
\label{sec:attribute share}
We have introduced graying and random rotation operations to erase the color distribution and the orientation accordingly. In Sec~\ref{method}, the two attributes are required to satisfy the second assumption, i.e. either different or connected to different semantic information between normal and anomalous data. We here argue that the second assumption is true for most cases in real-world scenarios. Taking the color as an example, Zhang~\emph{et al.}~\cite{zhang2016colorful} indicated that there are many statistical dependencies between the semantics and textures of grayscale images and their color versions. For each pixel, to restore its color information, the corresponding semantics and textures from the whole image should be taken into consideration. Even the color distribution is similar between normal and anomalous data, e.g. a yellow cat and a yellow car, the connected semantic information is different. Thus, when testing on anomalous data, the network is unable to utilize the unseen semantic information to restore the color properly. The same principle also applies to the orientation, which is connected with more complex semantics and textures. In conclusion, once the first assumption is satisfied, where AEM can erase attribute which connected to semantic information, the second assumption is easy to be satisfied as semantic information is likely to be different among different classes. Therefore, in future research, we do not need to be constrained to the choice of attributes but should consider more on how to extract more powerful semantic features under our attribute restoration framework.

\section{Conclusion and Future Work}
In this paper, we propose a novel technique named Attribute Restoration Network (ARNet) for anomaly detection. Attribute Erasing Module is employed to erase certain attributes. The ARNet is forced to learn the attribute related features to restore the original data. The restoration error is expected to be a good indicator of anomalous data. We experiment with two simple but effective attribute erasing operations: graying and random rotation, and show that our method not only outperforms state-of-the-art methods but also achieves high stability. Notably, there are still more operations to explore. These operations are likely to further improve the performance of ARNet for anomaly detection by effectively extracting semantic features. We look forward to the addition of more operations and the exploration of a more intelligent operations selection strategy. In addition, this way to learn semantic feature embeddings can also be applied to more fields, opening avenues for future research.

% references section------------------------------------------------------------------------------------------------------------------------------

\bibliographystyle{IEEEtran}
\bibliography{reference.bib}

\begin{thebibliography}{10}
\providecommand{\url}[1]{#1}
\csname url@samestyle\endcsname
\providecommand{\newblock}{\relax}
\providecommand{\bibinfo}[2]{#2}
\providecommand{\BIBentrySTDinterwordspacing}{\spaceskip=0pt\relax}
\providecommand{\BIBentryALTinterwordstretchfactor}{4}
\providecommand{\BIBentryALTinterwordspacing}{\spaceskip=\fontdimen2\font plus
\BIBentryALTinterwordstretchfactor\fontdimen3\font minus
  \fontdimen4\font\relax}
\providecommand{\BIBforeignlanguage}[2]{{%
\expandafter\ifx\csname l@#1\endcsname\relax
\typeout{** WARNING: IEEEtran.bst: No hyphenation pattern has been}%
\typeout{** loaded for the language `#1'. Using the pattern for}%
\typeout{** the default language instead.}%
\else
\language=\csname l@#1\endcsname
\fi
#2}}
\providecommand{\BIBdecl}{\relax}
\BIBdecl

\bibitem{chandola2009anomaly}
V.~Chandola, A.~Banerjee, and V.~Kumar, ``Anomaly detection: A survey,''
  \emph{Acm Computing Surveys}, vol.~41, no.~3, 2009.

\bibitem{ruff2018deep}
L.~Ruff, N.~G{\"o}rnitz, L.~Deecke, S.~A. Siddiqui, R.~Vandermeulen, A.~Binder,
  E.~M{\"u}ller, and M.~Kloft, ``Deep one-class classification,'' in
  \emph{International Conference on Machine Learning}, 2018.

\bibitem{schlegl2017unsupervised}
T.~Schlegl, P.~Seeb{\"o}ck, S.~M. Waldstein, U.~Schmidt-Erfurth, and G.~Langs,
  ``Unsupervised anomaly detection with generative adversarial networks to
  guide marker discovery,'' in \emph{International Conference on Information
  Processing in Medical Imaging}, 2017.

\bibitem{Akcay2018}
S.~Akcay, A.~Atapour-Abarghouei, and T.~P. Breckon, ``{GANomaly:
  Semi-Supervised Anomaly Detection via Adversarial Training},'' \emph{Asian
  Conference on Computer Vision}, 2018.

\bibitem{Sabokrou2018Adversarially}
M.~Sabokrou, M.~Khalooei, M.~Fathy, and E.~Adeli, ``Adversarially learned
  one-class classifier for novelty detection,'' in \emph{Proceedings of the
  IEEE international conference on computer vision}, 2018.

\bibitem{masci2011stacked}
J.~Masci, U.~Meier, D.~Cire{\c{s}}an, and J.~Schmidhuber, ``Stacked
  convolutional auto-encoders for hierarchical feature extraction,'' in
  \emph{International Conference on Artificial Neural Networks}, 2011.

\bibitem{Sakurada2014Anomaly}
M.~Sakurada and T.~Yairi, ``Anomaly detection using autoencoders with nonlinear
  dimensionality reduction,'' in \emph{Mlsda Workshop on Machine Learning for
  Sensory Data Analysis}, 2014.

\bibitem{russakovsky2015imagenet}
O.~Russakovsky, J.~Deng, H.~Su, J.~Krause, S.~Satheesh, S.~Ma, Z.~Huang,
  A.~Karpathy, A.~Khosla, M.~Bernstein \emph{et~al.}, ``Imagenet large scale
  visual recognition challenge,'' \emph{International Journal of Computer
  Vision}, 2015.

\bibitem{SimilarityMetricAutoencoding}
A.~B.~L. Larsen, S.~K. S{\o}nderby, H.~Larochelle, and O.~Winther,
  ``Autoencoding beyond pixels using a learned similarity metric,'' \emph{arXiv
  preprint arXiv:1512.09300}, 2015.

\bibitem{dosovitskiy2016generating}
A.~Dosovitskiy and T.~Brox, ``Generating images with perceptual similarity
  metrics based on deep networks,'' in \emph{Advances in neural information
  processing systems}, 2016, pp. 658--666.

\bibitem{gong2019memorizing}
D.~Gong, L.~Liu, V.~Le, B.~Saha, M.~R. Mansour, S.~Venkatesh, and A.~v.~d.
  Hengel, ``Memorizing normality to detect anomaly: Memory-augmented deep
  autoencoder for unsupervised anomaly detection,'' in \emph{Proceedings of the
  IEEE International Conference on Computer Vision}, 2019, pp. 1705--1714.

\bibitem{zong2018deep}
B.~Zong, Q.~Song, M.~R. Min, W.~Cheng, C.~Lumezanu, D.~Cho, and H.~Chen, ``Deep
  autoencoding gaussian mixture model for unsupervised anomaly detection,'' in
  \emph{International Conference on Learning Representations}, 2018.

\bibitem{deecke2018image}
L.~Deecke, R.~Vandermeulen, L.~Ruff, S.~Mandt, and M.~Kloft, ``Image anomaly
  detection with generative adversarial networks,'' in \emph{Joint European
  Conference on Machine Learning and Knowledge Discovery in Databases}.\hskip
  1em plus 0.5em minus 0.4em\relax Springer, 2018, pp. 3--17.

\bibitem{wang2019effective}
S.~Wang, Y.~Zeng, X.~Liu, E.~Zhu, J.~Yin, C.~Xu, and M.~Kloft, ``Effective
  end-to-end unsupervised outlier detection via inlier priority of
  discriminative network,'' in \emph{Advances in Neural Information Processing
  Systems}, 2019, pp. 5960--5973.

\bibitem{bergmann2020uninformed}
P.~Bergmann, M.~Fauser, D.~Sattlegger, and C.~Steger, ``Uninformed students:
  Student-teacher anomaly detection with discriminative latent embeddings,'' in
  \emph{Proceedings of the IEEE/CVF Conference on Computer Vision and Pattern
  Recognition}, 2020, pp. 4183--4192.

\bibitem{pathak2016context}
D.~Pathak, P.~Krahenbuhl, J.~Donahue, T.~Darrell, and A.~A. Efros, ``Context
  encoders: Feature learning by inpainting,'' in \emph{Proceedings of the IEEE
  conference on computer vision and pattern recognition}, 2016, pp. 2536--2544.

\bibitem{Jenni_2018_CVPR}
S.~Jenni and P.~Favaro, ``Self-supervised feature learning by learning to spot
  artifacts,'' in \emph{Proceedings of the IEEE international conference on
  computer vision}, June 2018.

\bibitem{zhang2016colorful}
R.~Zhang, P.~Isola, and A.~A. Efros, ``Colorful image colorization,'' in
  \emph{European conference on computer vision}.\hskip 1em plus 0.5em minus
  0.4em\relax Springer, 2016, pp. 649--666.

\bibitem{vincent2008extracting}
P.~Vincent, H.~Larochelle, Y.~Bengio, and P.-A. Manzagol, ``Extracting and
  composing robust features with denoising autoencoders,'' in \emph{Proceedings
  of the 25th international conference on Machine learning}, 2008, pp.
  1096--1103.

\bibitem{long2017towards}
Y.~Long, L.~Liu, and L.~Shao, ``Towards fine-grained open zero-shot learning:
  Inferring unseen visual features from attributes,'' in \emph{2017 IEEE Winter
  Conference on Applications of Computer Vision (WACV)}.\hskip 1em plus 0.5em
  minus 0.4em\relax IEEE, 2017, pp. 944--952.

\bibitem{bergmann2019mvtec}
P.~Bergmann, M.~Fauser, D.~Sattlegger, and C.~Steger, ``Mvtec ad--a
  comprehensive real-world dataset for unsupervised anomaly detection,'' in
  \emph{Proceedings of the IEEE international conference on computer vision},
  2019.

\bibitem{hendrycks2018benchmarking}
D.~Hendrycks and T.~G. Dietterich, ``Benchmarking neural network robustness to
  common corruptions and surface variations,'' \emph{International Conference
  on Learning Representations}, 2019.

\bibitem{maaten2008visualizing}
L.~v.~d. Maaten and G.~Hinton, ``Visualizing data using t-sne,'' \emph{Journal
  of machine learning research}, vol.~9, no. Nov, pp. 2579--2605, 2008.

\bibitem{Kiran2018An}
B.~R. Kiran, D.~M. Thomas, and R.~Parakkal, ``An overview of deep learning
  based methods for unsupervised and semi-supervised anomaly detection in
  videos,'' \emph{Journal of Imaging}, 2018.

\bibitem{chu2018sparse}
W.~Chu, H.~Xue, C.~Yao, and D.~Cai, ``Sparse coding guided spatiotemporal
  feature learning for abnormal event detection in large videos,'' \emph{IEEE
  Transactions on Multimedia}, vol.~21, no.~1, pp. 246--255, 2018.

\bibitem{xu2018anomaly}
K.~Xu, X.~Jiang, and T.~Sun, ``Anomaly detection based on stacked sparse coding
  with intraframe classification strategy,'' \emph{IEEE Transactions on
  Multimedia}, vol.~20, no.~5, pp. 1062--1074, 2018.

\bibitem{xu2019video}
K.~Xu, T.~Sun, and X.~Jiang, ``Video anomaly detection and localization based
  on an adaptive intra-frame classification network,'' \emph{IEEE Transactions
  on Multimedia}, 2019.

\bibitem{sabokrou2016video}
M.~Sabokrou, M.~Fathy, and M.~Hoseini, ``Video anomaly detection and
  localisation based on the sparsity and reconstruction error of
  auto-encoder,'' \emph{Electronics Letters}, vol.~52, no.~13, pp. 1122--1124,
  2016.

\bibitem{sabokrou2018avid}
M.~Sabokrou, M.~Pourreza, M.~Fayyaz, R.~Entezari, M.~Fathy, J.~Gall, and
  E.~Adeli, ``Avid: Adversarial visual irregularity detection,'' in \emph{Asian
  Conference on Computer Vision}.\hskip 1em plus 0.5em minus 0.4em\relax
  Springer, 2018, pp. 488--505.

\bibitem{sabokrou2020deep}
M.~Sabokrou, M.~Fathy, G.~Zhao, and E.~Adeli, ``Deep end-to-end one-class
  classifier,'' \emph{IEEE Transactions on Neural Networks and Learning
  Systems}, 2020.

\bibitem{sabokrou2018deep}
M.~Sabokrou, M.~Fayyaz, M.~Fathy, Z.~Moayed, and R.~Klette, ``Deep-anomaly:
  Fully convolutional neural network for fast anomaly detection in crowded
  scenes,'' \emph{Computer Vision and Image Understanding}, vol. 172, pp.
  88--97, 2018.

\bibitem{sabokrou2017deep}
M.~Sabokrou, M.~Fayyaz, M.~Fathy, and R.~Klette, ``Deep-cascade: Cascading 3d
  deep neural networks for fast anomaly detection and localization in crowded
  scenes,'' \emph{IEEE Transactions on Image Processing}, vol.~26, no.~4, pp.
  1992--2004, 2017.

\bibitem{Eskin2000Anomaly}
E.~Eskin, ``Anomaly detection over noisy data using learned probability
  distributions,'' in \emph{International Conference on Machine Learning},
  2000.

\bibitem{Yamanishi2000On}
K.~Yamanishi, J.~I. Takeuchi, G.~Williams, and P.~Milne, ``On-line unsupervised
  outlier detection using finite mixtures with discounting learning
  algorithms,'' \emph{Data Mining \& Knowledge Discovery}, 2000.

\bibitem{Rahmani2017Coherence}
M.~Rahmani and G.~K. Atia, ``Coherence pursuit: Fast, simple, and robust
  principal component analysis,'' \emph{IEEE Transactions on Signal
  Processing}, 2017.

\bibitem{Xu2012Robust}
H.~Xu, C.~Caramanis, and S.~Sanghavi, ``Robust pca via outlier pursuit,''
  \emph{IEEE Transactions on Information Theory}, 2012.

\bibitem{scholkopf2001estimating}
B.~Sch{\"o}lkopf, J.~C. Platt, J.~Shawe-Taylor, A.~J. Smola, and R.~C.
  Williamson, ``Estimating the support of a high-dimensional distribution,''
  \emph{Neural computation}, vol.~13, no.~7, pp. 1443--1471, 2001.

\bibitem{lee2017training}
K.~Lee, H.~Lee, K.~Lee, and J.~Shin, ``Training confidence-calibrated
  classifiers for detecting out-of-distribution samples,'' in
  \emph{International Conference on Learning Representations}, 2018.

\bibitem{hendrycks2018deep}
D.~Hendrycks, M.~Mazeika, and T.~Dietterich, ``Deep anomaly detection with
  outlier exposure,'' in \emph{International Conference on Learning
  Representations}, 2019.

\bibitem{an2015variational}
J.~An and S.~Cho, ``Variational autoencoder based anomaly detection using
  reconstruction probability,'' \emph{Special Lecture on IE}, 2015.

\bibitem{xia2015learning}
Y.~Xia, X.~Cao, F.~Wen, G.~Hua, and J.~Sun, ``Learning discriminative
  reconstructions for unsupervised outlier removal,'' in \emph{Proceedings of
  the IEEE international conference on computer vision}, 2015.

\bibitem{nicolau2016hybrid}
M.~Nicolau, J.~McDermott \emph{et~al.}, ``A hybrid autoencoder and density
  estimation model for anomaly detection,'' in \emph{International Conference
  on Parallel Problem Solving from Nature}.\hskip 1em plus 0.5em minus
  0.4em\relax Springer, 2016, pp. 717--726.

\bibitem{akccay2019skip}
S.~Akçay and A.~A.-A. andToby P.~Breckon, ``Skip-ganomaly: Skip connected and
  adversarially trained encoder-decoder anomaly detection,'' in \emph{2019
  International Joint Conference on Neural Networks (IJCNN)}, 2019.

\bibitem{wang2019advae}
X.~Wang, Y.~Du, S.~Lin, P.~Cui, Y.~Shen, and Y.~Yang, ``advae: A
  self-adversarial variational autoencoder with gaussian anomaly prior
  knowledge for anomaly detection,'' \emph{arXiv preprint arXiv:1903.00904},
  2019.

\bibitem{zenati2018efficient}
H.~Zenati, C.~S. Foo, B.~Lecouat, G.~Manek, and V.~R. Chandrasekhar,
  ``Efficient gan-based anomaly detection,'' \emph{arXiv preprint
  arXiv:1802.06222}, 2018.

\bibitem{perera2019ocgan}
P.~Perera, R.~Nallapati, and B.~Xiang, ``Ocgan: One-class novelty detection
  using gans with constrained latent representations,'' in \emph{Proceedings of
  the IEEE international conference on computer vision}, 2019.

\bibitem{golan2018deep}
I.~Golan and R.~El-Yaniv, ``Deep anomaly detection using geometric
  transformations,'' in \emph{Advances in Neural Information Processing
  Systems}, 2018.

\bibitem{gidaris2018unsupervised}
S.~Gidaris, P.~Singh, and N.~Komodakis, ``Unsupervised representation learning
  by predicting image rotations,'' in \emph{International Conference on
  Learning Representations}, 2019.

\bibitem{noroozi2016unsupervised}
M.~Noroozi and P.~Favaro, ``Unsupervised learning of visual representations by
  solving jigsaw puzzles,'' in \emph{European Conference on Computer
  Vision}.\hskip 1em plus 0.5em minus 0.4em\relax Springer, 2016, pp. 69--84.

\bibitem{doersch2015unsupervised}
C.~Doersch, A.~Gupta, and A.~A. Efros, ``Unsupervised visual representation
  learning by context prediction,'' in \emph{Proceedings of the IEEE
  international conference on computer vision}, 2015, pp. 1422--1430.

\bibitem{denton2016semi}
E.~Denton, S.~Gross, and R.~Fergus, ``Semi-supervised learning with
  context-conditional generative adversarial networks,'' \emph{arXiv preprint
  arXiv:1611.06430}, 2016.

\bibitem{kingma2013auto}
D.~P. Kingma and M.~Welling, ``Auto-encoding variational bayes,'' \emph{arXiv
  preprint arXiv:1312.6114}, 2013.

\bibitem{zhai2016deep}
S.~Zhai, Y.~Cheng, W.~Lu, and Z.~Zhang, ``Deep structured energy based models
  for anomaly detection,'' in \emph{International Conference on Machine
  Learning}, 2016, pp. 1100--1109.

\bibitem{lecun1998mnist}
Y.~LeCun, ``The mnist database of handwritten digits,'' \emph{http://yann.
  lecun. com/exdb/mnist/}, 1998.

\bibitem{xiao2017fashion}
H.~Xiao, K.~Rasul, and R.~Vollgraf, ``Fashion-mnist: a novel image dataset for
  benchmarking machine learning algorithms,'' \emph{arXiv preprint
  arXiv:1708.07747}, 2017.

\bibitem{krizhevsky2009learning}
A.~Krizhevsky and G.~Hinton, ``Learning multiple layers of features from tiny
  images,'' Citeseer, Tech. Rep., 2009.

\bibitem{blei2003latent}
D.~M. Blei, A.~Y. Ng, and M.~I. Jordan, ``Latent dirichlet allocation,''
  \emph{Journal of machine Learning research}, 2003.

\bibitem{unet}
O.~Ronneberger, P.~Fischer, and T.~Brox, ``U-net: Convolutional networks for
  biomedical image segmentation,'' in \emph{International Conference on Medical
  image computing and computer-assisted intervention}, 2015.

\bibitem{isola2017image}
P.~Isola, J.-Y. Zhu, T.~Zhou, and A.~A. Efros, ``Image-to-image translation
  with conditional adversarial networks,'' in \emph{Proceedings of the IEEE
  international conference on computer vision}, 2017.

\bibitem{bottou2010large}
L.~Bottou, ``Large-scale machine learning with stochastic gradient descent,''
  in \emph{Proceedings of COMPSTAT'2010}, 2010.

\bibitem{he2016deep}
K.~He, X.~Zhang, S.~Ren, and J.~Sun, ``Deep residual learning for image
  recognition,'' in \emph{Proceedings of the IEEE international conference on
  computer vision}, 2016.

\bibitem{huang2017densely}
G.~Huang, Z.~Liu, L.~Van Der~Maaten, and K.~Q. Weinberger, ``Densely connected
  convolutional networks.'' in \emph{Proceedings of the IEEE international
  conference on computer vision}, 2017.

\bibitem{luo2017revisit}
W.~Luo, W.~Liu, and S.~Gao, ``A revisit of sparse coding based anomaly
  detection in stacked rnn framework,'' in \emph{Proceedings of the IEEE
  international conference on computer vision}, 2017.

\bibitem{zhao2017spatio}
Y.~Zhao, B.~Deng, C.~Shen, Y.~Liu, H.~Lu, and X.-S. Hua, ``Spatio-temporal
  autoencoder for video anomaly detection,'' in \emph{Proceedings of the 25th
  ACM international conference on Multimedia}.\hskip 1em plus 0.5em minus
  0.4em\relax ACM, 2017.

\bibitem{hasan2016learning}
M.~Hasan, J.~Choi, J.~Neumann, A.~K. Roy-Chowdhury, and L.~S. Davis, ``Learning
  temporal regularity in video sequences,'' in \emph{Proceedings of the IEEE
  international conference on computer vision}, 2016.

\end{thebibliography}

\appendices

\section{Index for ImageNet~\cite{russakovsky2015imagenet}}
We group categories into 10 superclasses by merging similar categories for anomaly detection. Table~\ref{tal:index1} shows the specific category index.

\renewcommand \arraystretch{0.9}
\begin{table}[!htb]
  \centering
  \caption{Index of clustering results for ImageNet.}
  \label{tal:index1}
  \footnotesize
  \begin{tabular}{cx{1.5cm}x{5cm}}
    \toprule
		$c_i$ & Label & Index \\
		\cmidrule(lr){1-1} \cmidrule(lr){2-2} \cmidrule(lr){3-3} 
		0 & Snake & n01728920, n01728572, n01729322, \\&& n01734418, n01737021, n01740131, \\&& n01735189\\
		1 & Finch & n01530575, n01531178, n01532829,  n01534433, n01795545, n01796340\\
		2 & Spider & n01773157, n01773549, n01774384,  n01775062, n01773797, n01774750\\
		3 & Big cat & n02128385, n02128925, n02129604,   n02130308, n02128757, n02129165\\
	    4 & Beetle & n02165105, n02165456, n02169497, n02177972, n02167151\\
		5 & Wading bird & n02007558, n02012849, n02013706,  n02018795, n02006656\\
		6 & Monkey & n02486261, n02486410, n02488291,  n02489166\\
		7 & Fungus & n12985857, n13037406, n13054560,  n13040303\\
		8 & Cat & n02123045, n02123394, n02124075,   n02123159\\
		9 & Dog & n02088364, n02105412, n02106030, n02106166, n02106662,  n02106550,  n02088466, n02093754, n02091635\\
        \bottomrule
		\end{tabular}
\end{table}

\section{Index for MVTec AD~\cite{bergmann2019mvtec}}
MVTec AD dataset contains 5354 high-resolution color images of different object and texture categories. It contains normal images intended for training and images with anomalies intended for testing. The anomalies manifest themselves in the form of over 70 different types of defects such as scratches, dents, contaminations, and various structural changes. Table~\ref{tal:index2} shows class names and anomalous types for each categories.

\renewcommand \arraystretch{0.9}
\begin{table}[!htb]
  \centering
  \caption{Class names and anomalous types of MVTec AD.}
  \footnotesize
  \begin{tabular}{cx{1.4cm}x{5.6cm}}
    \toprule
		$c_i$ & Class Name & Anomalous Types\\
		\cmidrule(lr){1-1} \cmidrule(lr){2-2} \cmidrule(lr){3-3} 
		0 & Bottle & broken large, broken small, contamination\\
		1 & Capsule & crack, faulty imprint, poke, scratch, squeeze\\
		2 & Grid & bent, broken, glue, metal contamination, thread\\
		3 & Leather & color, cut, fold, glue, poke\\
	    4 & Pill & color, combined, contamination, crack, faulty imprint, pill type, scratch\\
		5 & Tile & crack, glue strip, gray stroke, oil, rough\\
		6 & Transistor & bent, cut, damaged, misplaced\\
		7 & Zipper & broken teeth, combined, fabric border, fabric interior, rough, split teeth, squeezed teeth\\
		8 & Cable & bent wire, cable swap, combined, cut inner insulation, cut outer insulation, missing cable, missing wire, poke insulation\\
		9 & Carpet & color, cut, hole, metal contamination, thread\\
		10 & Hazelnut & crack, cut, hole, print\\
	    11 & Metal nut & bent, color, flip, scratch\\
		12 & Screw & manipulated front, scratch head, scratch neck, thread side, thread top\\
		13 & Toothbrush & defective\\
	    14 & Wood & color, combined, hole, liquid, scratch\\
        \bottomrule
		\end{tabular}
	\label{tal:index2}
\end{table}

\renewcommand \arraystretch{0.9}
\begin{table}[tbph]
  \centering
  \footnotesize
  \caption{Structure of ARNet.}
  \label{tal:ARNet}
     \begin{tabular}{ccc}
      \toprule
      Layer & Input & Output\\
      \cmidrule(lr){1-1}\cmidrule(lr){2-2}\cmidrule(lr){3-3}
      $3\times 3\times 64$ & $x\ (1 \times H \times W)$ & $x_{0-1}\ (64 \times H \times W)$\\
      $3\times 3\times 64$ & $x_{0-1}$ & $x_{0-2}\ (64 \times H \times W)$\\ 
      MaxPool & $x_{0-2}$ & $x_{1-1}\ (64 \times 1/2 H \times 1/2W)$ \\
      $3\times 3\times 128$ & $x_{1-1}$ & $x_{1-2}\ (128 \times 1/2 H \times 1/2W)$\\ 
      $3\times 3\times 128$ & $x_{1-2}$ & $x_{1-3}\ (128 \times 1/2 H \times 1/2W)$\\ 
      MaxPool & $x_{1-3}$ & $x_{2-1}\ (128 \times 1/4 H \times 1/4W)$ \\
      $3\times 3\times 256$ & $x_{2-1}$ & $x_{2-2}\ (256 \times 1/4 H \times 1/4W)$\\ 
      $3\times 3\times 256$ & $x_{2-2}$ & $x_{2-3}\ (256 \times 1/4 H \times 1/4W)$\\ 
      MaxPool & $x_{2-3}$ & $x_{3-1}\ (256 \times 1/8 H \times 1/8W)$ \\
      $3\times 3\times 512$ & $x_{3-1}$ & $x_{3-2}\ (256 \times 1/8 H \times 1/8W)$\\ 
      $3\times 3\times 512$ & $x_{3-2}$ & $x_{3-3}\ (256 \times 1/8 H \times 1/8W)$\\ 
      MaxPool & $x_{3-3}$ & $x_{4-1}\ (256 \times 1/8 H \times 1/16W)$ \\
      $3\times 3\times 512$ & $x_{4-1}$ & $x_{4-2}\ (512 \times 1/16 H \times 1/16W)$\\ 
      $3\times 3\times 512$ & $x_{4-2}$ & $x_{4-3}\ (512 \times 1/16 H \times 1/16W)$\\ 
      UpSample & $x_{4-3}$ & $up_{3-1}\ (512 \times 1/8 H \times 1/8W)$ \\
      $3\times 3\times 256$ & $[up_{3-1},x_{3-3}]$ & $up_{3-2}\ (256 \times 1/8 H \times 1/8W)$\\ 
      $3\times 3\times 256$ & $up_{3-2}$ & $up_{3-3}\ (256 \times 1/8 H \times 1/8W)$\\ 
      UpSample & $up_{3-3}$ & $up_{2-1}\ (256 \times 1/4 H \times 1/4W)$ \\
      $3\times 3\times 128$ & $[up_{2-1},x_{2-3}]$ & $up_{2-2}\ (128 \times 1/4 H \times 1/4W)$\\ 
      $3\times 3\times 128$ & $up_{2-2}$ & $up_{2-3}\ (128 \times 1/4 H \times 1/4W)$\\ 
      UpSample & $up_{2-3}$ & $up_{1-1}\ (128 \times 1/2 H \times 1/2W)$ \\
      $3\times 3\times 64$ & $[up_{1-1},x_{1-3}]$ & $up_{1-2}\ (64 \times 1/2 H \times 1/2W)$\\ 
      $3\times 3\times 64$ & $up_{1-2}$ & $up_{1-3}\ (64 \times 1/2 H \times 1/2W)$\\ 
      UpSample & $x_{1-3}$ & $up_{0-1}\ (64 \times H \times W)$ \\
      $3\times 3\times 64$ & $[up_{0-1},x_{0-2}]$ & $up_{0-2}\ (64 \times H \times W)$\\ 
      $3\times 3\times 64$ & $up_{0-2}$ & $up_{0-3}\ (64 \times H \times W)$\\ 

      $3\times 3\times 3$ & $up_{0-3}$ & $output\ (3 \times H \times W)$\\ 
      \bottomrule
    \end{tabular}
\end{table}

\begin{figure}[!h]
\centering
\begin{minipage}[t]{0.24\textwidth}
\centering
\includegraphics[width=4.3cm]{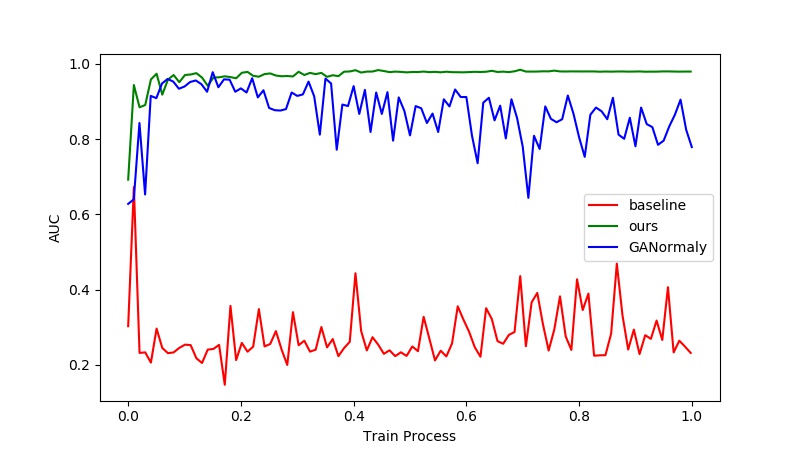}\\
(a) Class 0
%\caption{World Map}
\end{minipage}
\begin{minipage}[t]{0.24\textwidth}
\centering
\includegraphics[width=4.3cm]{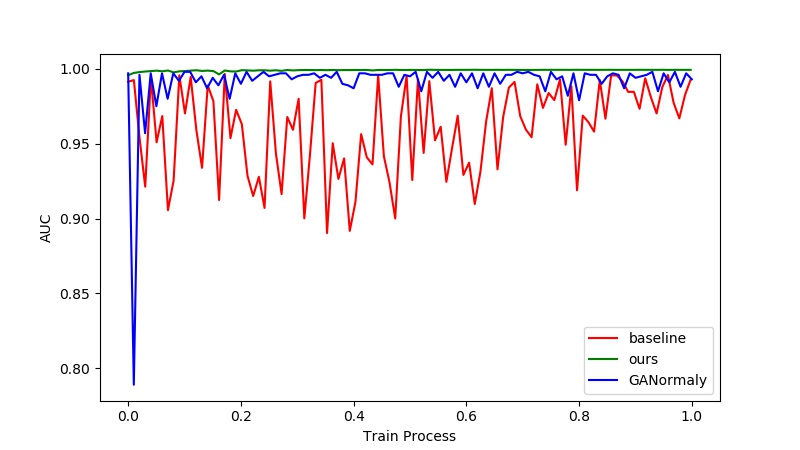}\\
(b) Class 1
%\caption{Concrete and Constructions}
\end{minipage}
\begin{minipage}[t]{0.24\textwidth}
\centering
\includegraphics[width=4.3cm]{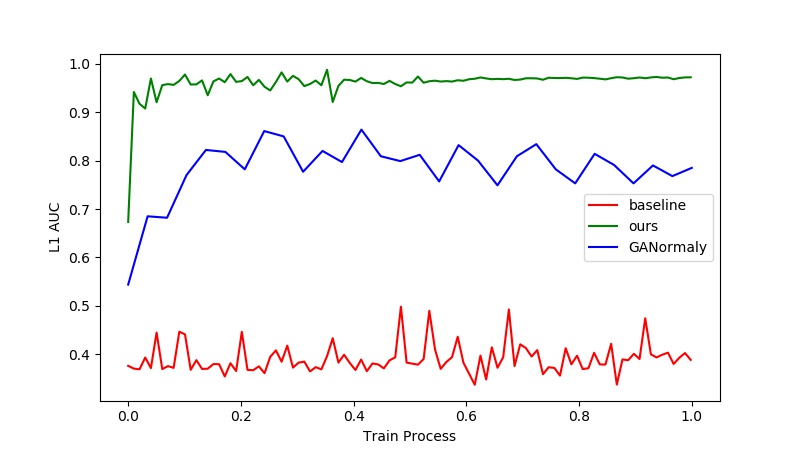}\\
(c) Class 2
%\caption{World Map}
\end{minipage}
\begin{minipage}[t]{0.24\textwidth}
\centering
\includegraphics[width=4.3cm]{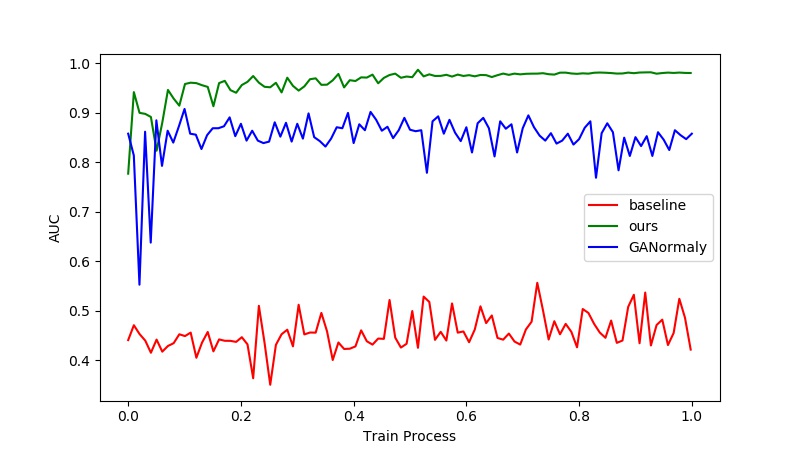}\\
(d) Class 3
%\caption{Concrete and Constructions}
\end{minipage}
\begin{minipage}[t]{0.24\textwidth}
\centering
\includegraphics[width=4.3cm]{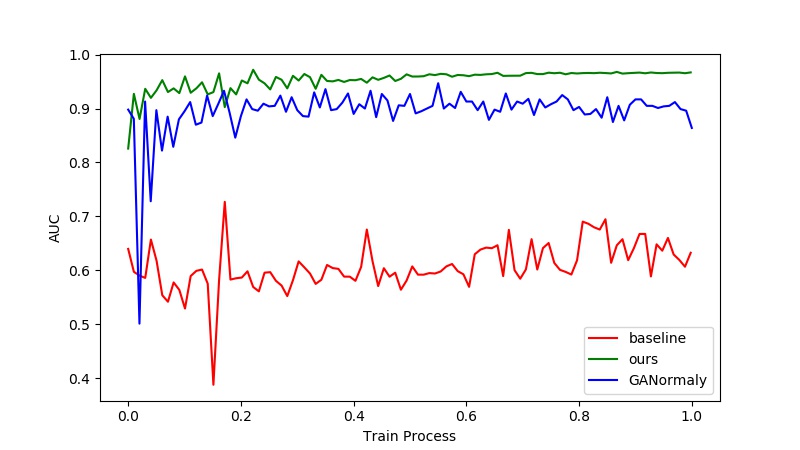}\\
(e) Class 4
%\caption{World Map}
\end{minipage}
\begin{minipage}[t]{0.24\textwidth}
\centering
\includegraphics[width=4.3cm]{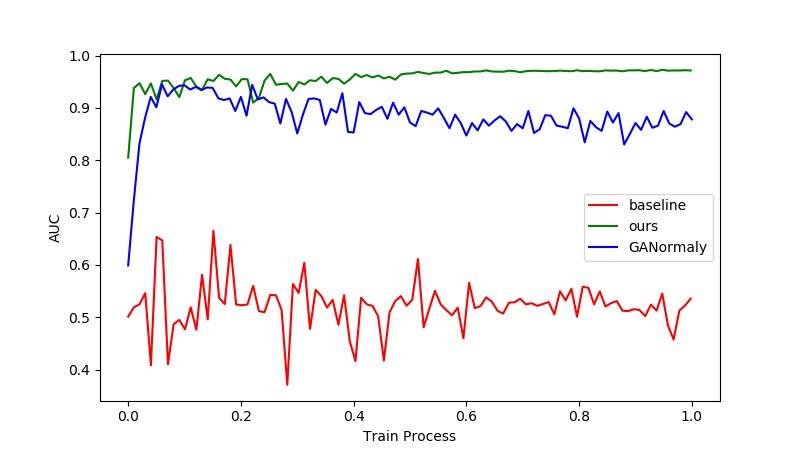}\\
(f) Class 5
%\caption{Concrete and Constructions}
\end{minipage}
\begin{minipage}[t]{0.24\textwidth}
\centering
\includegraphics[width=4.3cm]{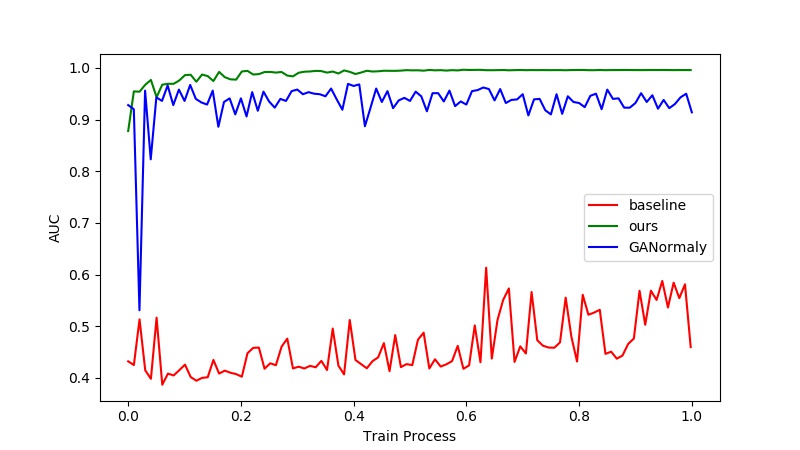}\\
(g) Class 6
%\caption{World Map}
\end{minipage}
\begin{minipage}[t]{0.24\textwidth}
\centering
\includegraphics[width=4.3cm]{img/train_log/L1_auc_mnist_7.jpg}\\
(h) Class 7
%\caption{Concrete and Constructions}
\end{minipage}
\begin{minipage}[t]{0.24\textwidth}
\centering
\includegraphics[width=4.3cm]{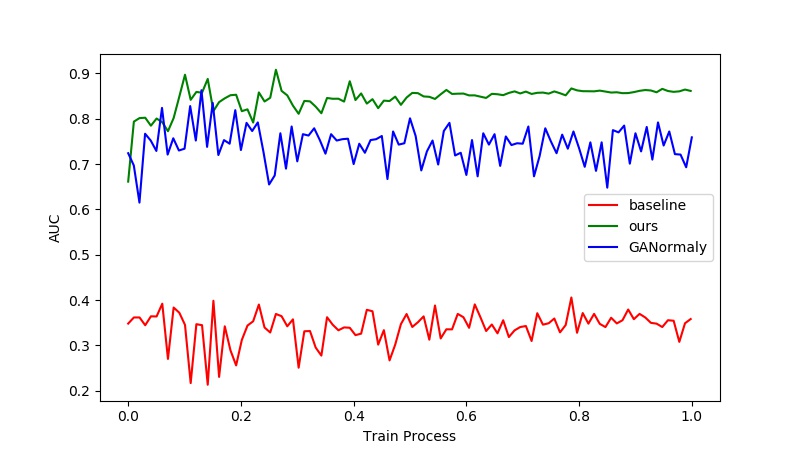}\\
(i) Class 8
%\caption{World Map}
\end{minipage}
\begin{minipage}[t]{0.24\textwidth}
\centering
\includegraphics[width=4.3cm]{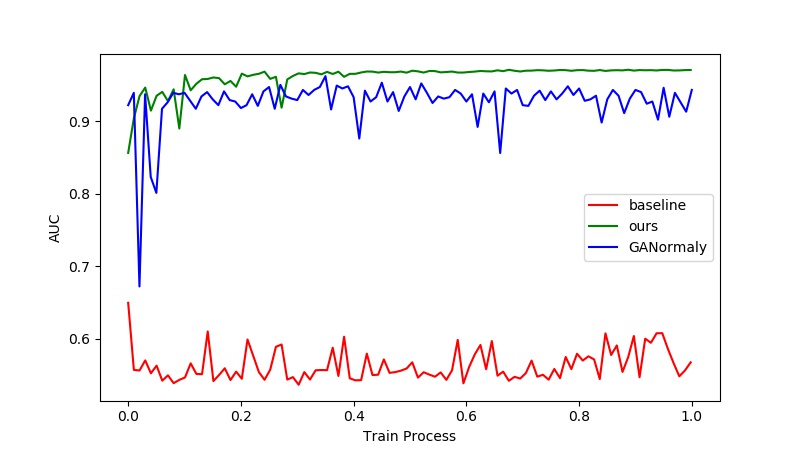}\\
(j) Class 9
%\caption{Concrete and Constructions}
\end{minipage}
\caption{Reported Accuracy under the $L_1$ metric on the test dataset of MNIST\cite{lecun1998mnist}. Ten sub-images represent the cases where the digit ``0''-``9'' is set as the normal category by order.}
\label{fig:stable3}
\end{figure}

\section{Model Structure of ARNet}
Table~\ref{tal:ARNet} shows the model structure of ARNet. It bases on an encoder-decoder framework. It totally has 4 blocks for the encoder and 4 blocks for the decoder. Each block has a maxpooling or an upsampling operation, following two $3\times 3$ convolutional layers. Skip-connection operations are added to facilitate the backpropagation of the gradient and improve the performance of image restoration. 
%Classification branch mentioned in the original paper contains two residual blocks. Each residual block contains two convolutional layers, one of which has a stride of 2. The residual part uses a $3\times 3$ convolutional layer with stride 2 for down-sampling. Finally, a global average pooling layer and a FC layer are used as classifier. The supervised information is its own type of transformation.

\begin{figure}[htbp]
\centering
\begin{minipage}[t]{0.24\textwidth}
\centering
\includegraphics[width=4.3cm]{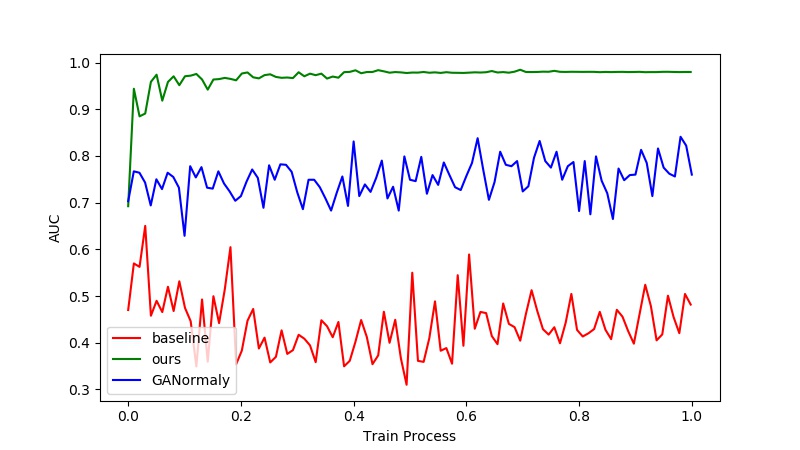}\\
(a) Class 0
%\caption{World Map}
\end{minipage}
\begin{minipage}[t]{0.24\textwidth}
\centering
\includegraphics[width=4.3cm]{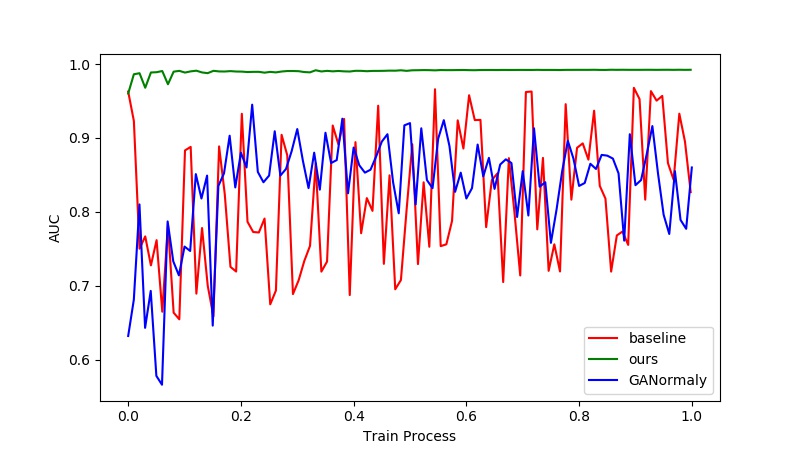}\\
(b) Class 1
%\caption{Concrete and Constructions}
\end{minipage}
\begin{minipage}[t]{0.24\textwidth}
\centering
\includegraphics[width=4.3cm]{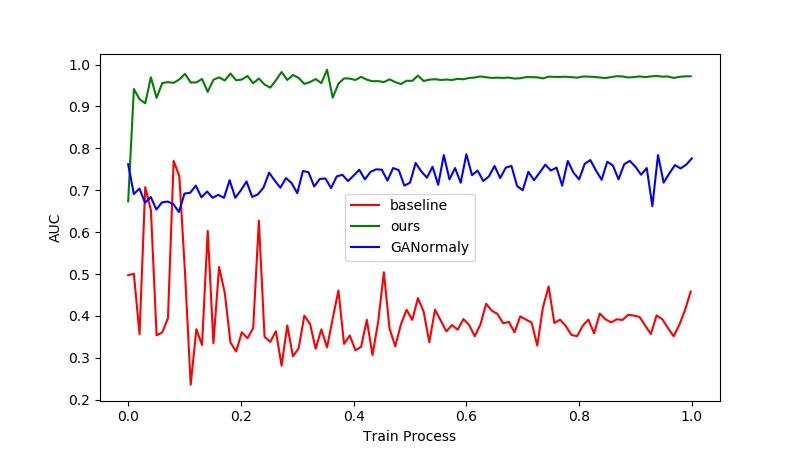}\\
(c) Class 2
\end{minipage}
\begin{minipage}[t]{0.24\textwidth}
\centering
\includegraphics[width=4.3cm]{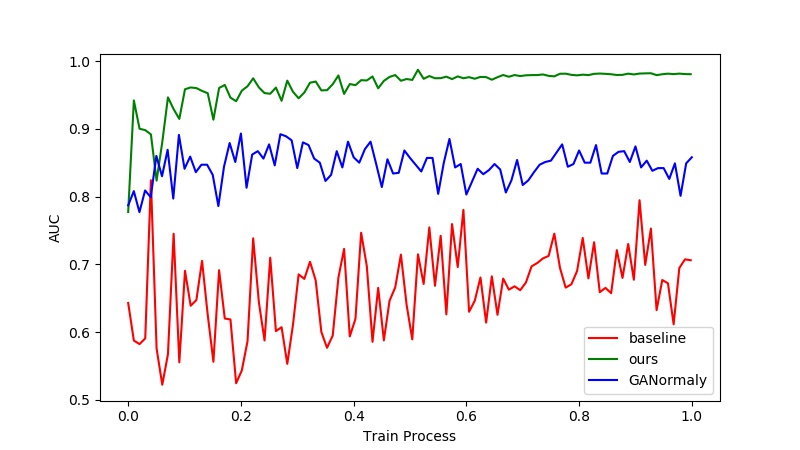}\\
(d) Class 3
\end{minipage}
\begin{minipage}[t]{0.24\textwidth}
\centering
\includegraphics[width=4.3cm]{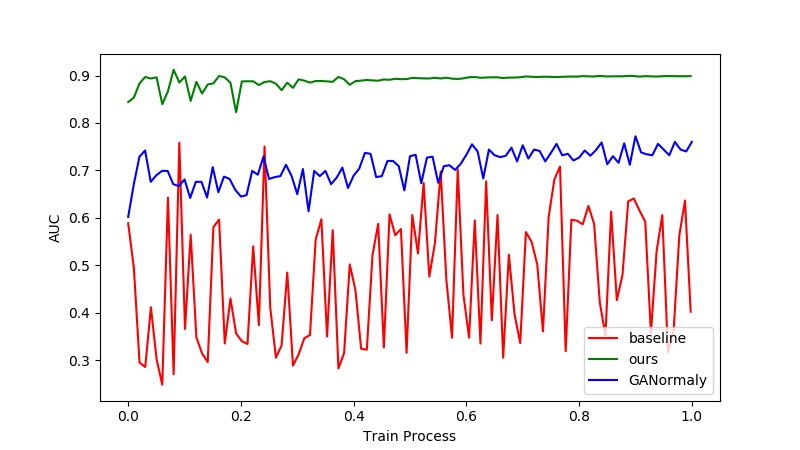}\\
(e) Class 4
\end{minipage}
\begin{minipage}[t]{0.24\textwidth}
\centering
\includegraphics[width=4.3cm]{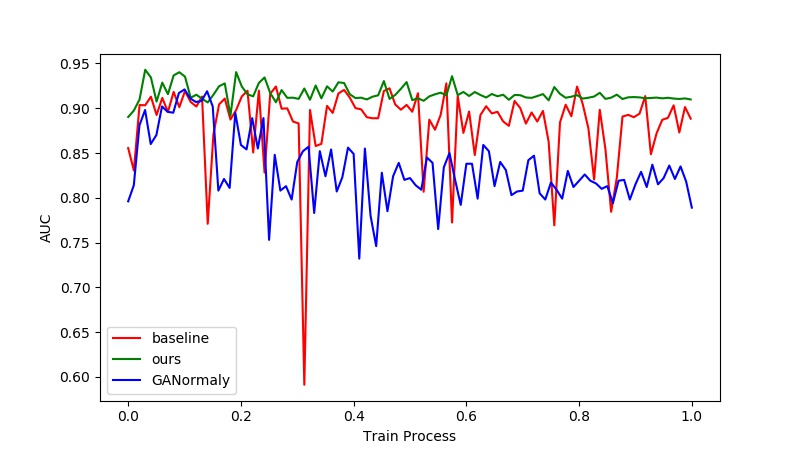}\\
(f) Class 5
\end{minipage}
\begin{minipage}[t]{0.24\textwidth}
\centering
\includegraphics[width=4.3cm]{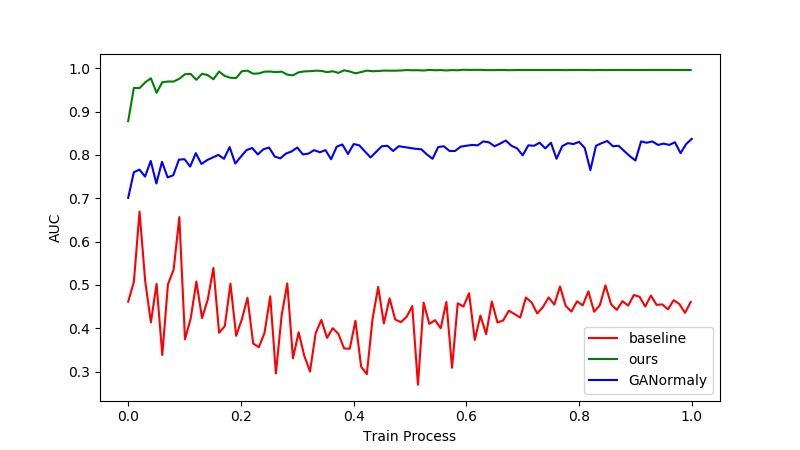}\\
(g) Class 6
\end{minipage}
\begin{minipage}[t]{0.24\textwidth}
\centering
\includegraphics[width=4.3cm]{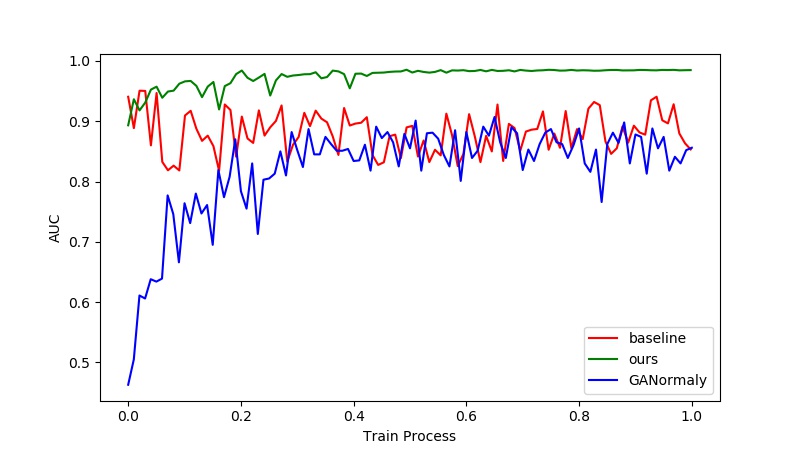}\\
(h) Class 7
\end{minipage}
\begin{minipage}[t]{0.24\textwidth}
\centering
\includegraphics[width=4.3cm]{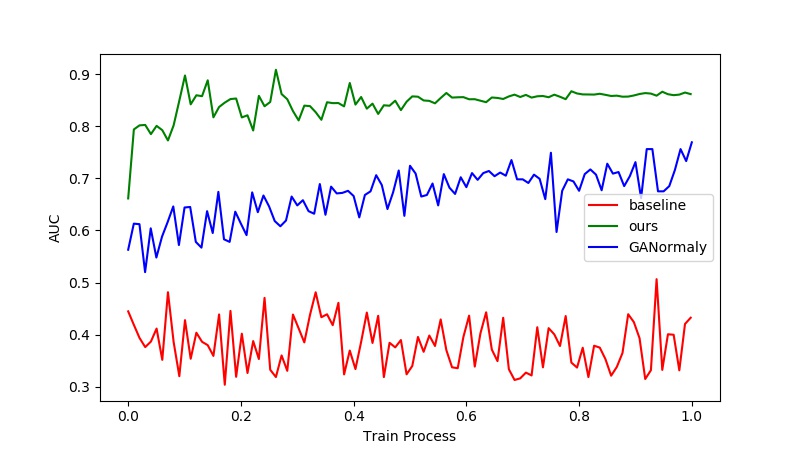}\\
(i) Class 8
\end{minipage}
\begin{minipage}[t]{0.24\textwidth}
\centering
\includegraphics[width=4.3cm]{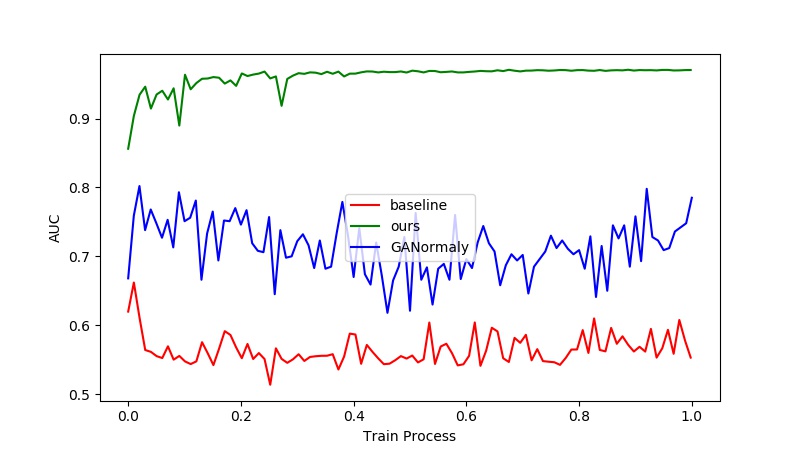}\\
(j) Class 9
\end{minipage}
\caption{Reported Accuracy under the $L_1$ metric on the test dataset of Fashion-MNIST\cite{xiao2017fashion}. Ten sub-images represent the cases where the class 0 - 9 is set as the normal category by order.}
\label{fig:stable2}
\end{figure}

\renewcommand \arraystretch{0.95}
\begin{table}[!htb]
	\centering
	\caption{Average area under the ROC curve (AUROC) in \% of anomaly detection methods on ShanghaiTech~\cite{luo2017revisit} dataset. The best performing method in each experiment is in bold.}
	\small
	\begin{tabular}{cx{3cm}x{1.1cm}}
		\toprule
		Methods & Temporal Dependency? & AUROC \\
		\cmidrule(lr){1-1} \cmidrule(lr){2-2} \cmidrule(lr){3-3}
		TSC~\cite{luo2017revisit} & \checkmark & 67.9\\
		StackRNN~\cite{luo2017revisit} & \checkmark & 68.0\\
		AE-Conv3D~\cite{zhao2017spatio} & \checkmark & 69.7\\
		MemAE~\cite{gong2019memorizing} & \checkmark & 71.2\\
		\cmidrule(lr){1-1} \cmidrule(lr){2-2} \cmidrule(lr){3-3}
		AE-Conv2D~\cite{hasan2016learning} & \ding{55} & 60.9\\
		OURS & \ding{55} & \textbf{72.5}\\
		\bottomrule
		\end{tabular}
	\label{tal:shanghai}
\end{table}

\begin{figure}[!h]
  \begin{minipage}[t]{0.114\textwidth}
\centering
\includegraphics[width=2cm]{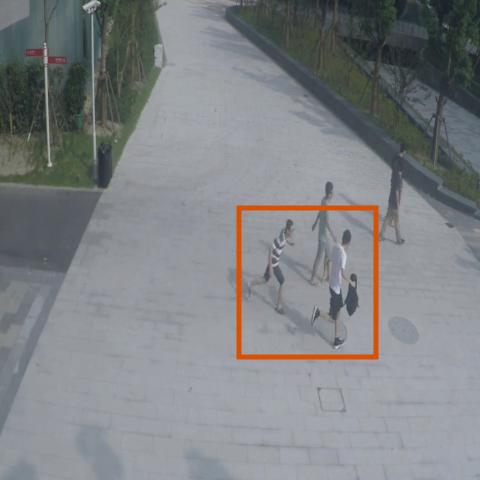}
\footnotesize
(a) Frame
\end{minipage}
\begin{minipage}[t]{0.114\textwidth}
\centering
\includegraphics[width=2cm]{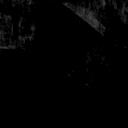}
\footnotesize
(b) AE-Conv2D
\end{minipage}
\begin{minipage}[t]{0.114\textwidth}
\centering
\includegraphics[width=2cm]{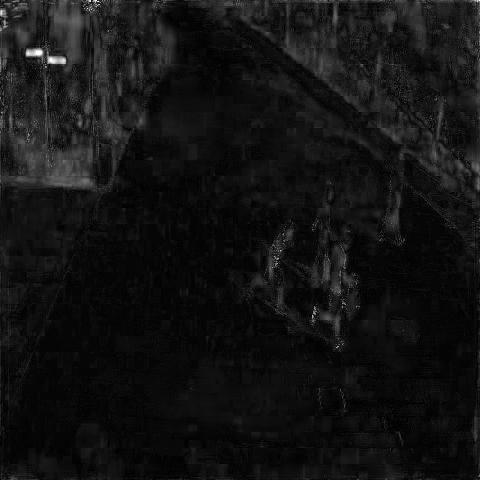}
\footnotesize
(c) ARNet(G)
\end{minipage}
\begin{minipage}[t]{0.114\textwidth}
\centering
\includegraphics[width=2cm]{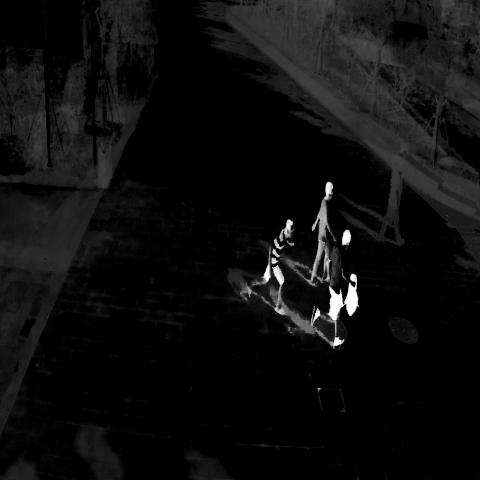}
\footnotesize
(d) ARNet(G+R)
\end{minipage}
\caption{Restoration error maps of AE and ARNet on an anomalous frame of ShanghaiTech. Chasing is the anomalous event in this frame (red bounding box). ``G'' means graying and ``R'' means random rotation transformation. ARNet can significantly highlight the anomalous parts in the scene.}
  \label{fig:shanghaitech}
\end{figure}

\section{Model Stability}
We argue that our proposed method achieves more robust performance. The main challenge in the task of anomaly detection is the lack of negative samples. Without validation, model stability tends to be more important than traditional data classification tasks. We train three models, including ARNet, traditional autoencoder~\cite{masci2011stacked} and GANomaly~\cite{Akcay2018}, respectively on each category of MNIST~\cite{lecun1998mnist} and Fashion-MNIST~\cite{xiao2017fashion} datasets and test models every 5 epochs along with training. The traditional autoencoder~\cite{masci2011stacked} and GANomaly~\cite{Akcay2018} are set as our baseline model. The model performance of validation during the training process is shown in Figure~\ref{fig:stable3} and Figure~\ref{fig:stable2}, from which we can see the performance of our ARNet method always converges in a high position; moreover, ARNet shows the highest performance stability at the end of the training process.

\section{Experiments on Video Anomaly Detection}

Video anomaly detection, which is distinguished from image-level anomaly detection, requires detections of anomalous objects and strenuous motions in the video data. We here experiment on a most recent video anomaly detection benchmark dataset ShanghaiTech~\cite{luo2017revisit}, comparing our methods with other state-of-the-arts.

\noindent\textbf{ShanghaiTech.} ShanghaiTech~\cite{luo2017revisit} has $13$ scenes with complex light conditions and camera angles. It contains $130$ anomalous events and over $270,000$ training frames. In the dataset, objects except for pedestrians (\emph{e.g.},~vehicles) and strenuous motion (\emph{e.g.},~fighting and chasing) are treated as anomalies. 

\noindent\textbf{Comparison with state-of-the-art methods.} Since our ARNet is designed for image-level anomaly detection, different from some state-of-the-arts~\cite{luo2017revisit,zhao2017spatio,gong2019memorizing}, we use single frames but not stacking neighbor frames as inputs. In order to apply the random rotation transformation, we resize all the images into $480\times 480$. We here use ResNet34~\cite{he2016deep} as our encoder. Following~\cite{hasan2016learning,luo2017revisit,gong2019memorizing}, we obtain the normality score $p_u$ of the $u$th frame by normalizing the errors to range $[0, 1]$:

\begin{equation}
    p_u = 1 - \frac{e_u-\min_u(e_u)}{\max_u(e_u)-\min_u(e_u)},
\end{equation}
where $e_u$ denotes the restoration error of the $u$th frame in a video episode. The value of $p_u$ closer to $0$ indicates the frame is more likely an anomalous frame. Table~\ref{tal:shanghai} shows the AUROC values on ShanghaiTech dataset. Results show that our ARNet outperforms all the state-of-the-arts, including some temporal dependent methods~\cite{luo2017revisit,zhao2017spatio,gong2019memorizing}.

\noindent\textbf{Visualization Analysis.} Figure~\ref{fig:shanghaitech} shows restoration error maps of AE and ARNet on an anomalous frame of ShanghaiTech, in which the highlight regions (regions with high restoration error) are considered as anomalous. In this frame, human chasing is the anomalous event (red bounding box in Figure~\ref{fig:shanghaitech} (a)). Due to good model generalization, AE reconstructs this frame properly even including the anomalous event (human chasing), leading to a low reconstruction error (reconstruction error map almost all black in Figure~\ref{fig:shanghaitech} (b)) Thus, AE cannot correctly detect this anomalous event. On the contrary, ARNet can not restore the anomalous region properly and significantly highlights the anomalous regions in the restoration error maps in Figure~\ref{fig:shanghaitech} (c and d). This is the reason why ARNet outperforms state-of-the-arts in video anomaly detection.

\ifCLASSOPTIONcaptionsoff
  \newpage
\fi
\end{document}